\documentclass[11pt]{article}

\usepackage{array}
\newcolumntype{P}[1]{>{\centering\arraybackslash}p{#1}}

\usepackage{booktabs} % move this to preamble and uncomment

\usepackage{hyperref}

\usepackage{tikz}
\usepackage{verbatim}

\usepackage{footnote}

\usepackage{bbm}

\usepackage{amssymb}
\usepackage{lineno}
\usepackage{mathtools}
\usepackage{graphicx,url}
\usepackage{graphics}

\DeclareMathOperator*{\E}{\mathbb{E}}
\DeclareMathOperator*{\R}{\mathcal{R}}

\DeclareMathOperator*{\I}{\mathbbm{1}}
\DeclareMathOperator*{\F}{\mathcal{F}}

\DeclareMathOperator*{\argmin}{argmin}

\usepackage{setspace,graphicx,epstopdf,amsmath,amsfonts,amssymb,amsthm,versionPO}
\usepackage{marginnote,datetime,enumitem,rotating,fancyvrb,scalerel}
\usepackage{hyperref}
\excludeversion{notes}		% Include notes?
\includeversion{links}          % Turn hyperlinks on?

\iflinks{}{\hypersetup{draft=true}}

\ifnotes{%
\usepackage[margin=1in,paperwidth=10in,right=2.5in]{geometry}%
\usepackage[textwidth=1.4in,shadow,colorinlistoftodos]{todonotes}%
}{%
\usepackage[margin=1in]{geometry}%
\usepackage[disable]{todonotes}%
}

\usepackage{color,soul}
\usepackage{xcolor}
\usepackage{framed}
\colorlet{shadecolor}{yellow!20}

\makeatletter\let\chapter\@undefined\makeatother % Undefine \chapter for todonotes

\usepackage{indentfirst} % Indent first sentence of a new section.
\usepackage{jfe}          % JFE-specific formatting of sections, etc.

\newtheorem{proposition}{Proposition}

\newtheorem{definition}{Definition}

\usepackage[
  backend=biber,
  style=numeric, % default
]{biblatex}
\usepackage{float}
\usepackage[caption = false]{subfig}

\begin{document}

\setlist{noitemsep}  % Reduce space between list items (itemize, enumerate, etc.)
%\onehalfspacing      % Use 1.5 spacing
% Use endnotes instead of footnotes - redefine \footnote command
% Old title: Theory-based deep residual neural networks: A synergy of decision-making theories and deep neural networks

\title{\textbf{Theory-based residual neural networks: A synergy of discrete choice models and deep neural networks}}

\author{Shenhao Wang\footnote{Corresponding author: shenhao@mit.edu} \\
  Baichuan Mo \\
  Jinhua Zhao \\
  \\
  \small{Massachusetts Institute of Technology} \\
  \small{Cambridge, MA 02139} \\
  \small{Oct, 2020} \\
}

\date{}              % No date for final submission

% Create title page with no page number
\renewcommand{\thefootnote}{\fnsymbol{footnote}}

\singlespacing
\maketitle

\vspace{-.2in}
\begin{abstract}
\noindent Researchers often treat data-driven and theory-driven models as two disparate or even conflicting methods in travel behavior analysis. However, the two methods are highly complementary because data-driven methods are more predictive but less interpretable and robust, while theory-driven methods are more interpretable and robust but less predictive. Using their complementary nature, this study designs a theory-based residual neural network (TB-ResNet) framework, which synergizes discrete choice models (DCMs) and deep neural networks (DNNs) based on their shared utility interpretation. The TB-ResNet framework is simple, as it uses a ($\delta$, 1-$\delta$) weighting to take advantage of DCMs' simplicity and DNNs' richness, and to prevent underfitting from the DCMs and overfitting from the DNNs. This framework is also flexible: three instances of TB-ResNets are designed based on multinomial logit model (MNL-ResNets), prospect theory (PT-ResNets), and hyperbolic discounting (HD-ResNets), which are tested on three data sets. Compared to pure DCMs, the TB-ResNets provide greater prediction accuracy and reveal a richer set of behavioral mechanisms owing to the utility function augmented by the DNN component in the TB-ResNets. Compared to pure DNNs, the TB-ResNets can modestly improve prediction and significantly improve interpretation and robustness, because the DCM component in the TB-ResNets stabilizes the utility functions and input gradients. Overall, this study demonstrates that it is both feasible and desirable to synergize DCMs and DNNs by combining their utility specifications under a TB-ResNet framework. Although some limitations remain, this TB-ResNet framework is an important first step to create mutual benefits between DCMs and DNNs for travel behavior modeling, with joint improvement in prediction, interpretation, and robustness.
\end{abstract}
\medskip
%\noindent \textit{JEL classification}: XXX, YYY.
%\medskip
%\textit{Keywords}: \LaTeX; papers with no content.

\thispagestyle{empty}

\clearpage

\onehalfspacing
\setcounter{footnote}{0}
\renewcommand{\thefootnote}{\arabic{footnote}}
\setcounter{page}{1}

\section{Introduction}
\label{s:1}
\noindent
As machine learning (ML) is increasingly used in the transportation field, we observe a tension between data-driven ML methods and classical theory-driven methods. Take travel behavior research as an example: researchers can analyze travel mode choice by using discrete choice models (DCMs) under the framework of random utility maximization (RUM), or using data-driven methods such as ML classifiers without any substantial behavioral understanding. This tension creates practical difficulty in choosing one method over the other, and prevents scholars from tackling travel behavior problems under a unified framework. This tension even delineates a form of partisan line within the transportation research community: researchers using data-driven methods focus on computational perspectives and prediction accuracy, while researchers using theory-driven methods focus on interpretation, economic information, and behavioral foundations. 

However, a closer examination reveals that the two methods are complementary in terms of prediction, interpretation, and robustness, prompting us to ask how to synergize them rather than treating them as disparate or even conflicting methods. As summarized in Table 1, deep neural networks (DNNs) and DCMs can be complementary because the former are more predictive, but less interpretable and robust, while the latter are less predictive, but more interpretable and robust. While DNNs are widely known as highly predictive \cite{Kotsiantis2007, Karlaftis2011, Krizhevsky2012, LeCun2015, Glaeser2018}, researchers often contend that DNNs lack interpretability \cite{Kotsiantis2007, Lipton2016, Boshi_Velez2017}, which is crucial in analyzing individual behavior for reasons such as safety in autonomous vehicles, knowledge distillation in research, and transparency in governance \cite{Lipton2016, Boshi_Velez2017, Freitas2014}. DNNs are also found to lack robustness, creating a brittle system vulnerable to small random noises or adversarial attacks. On the other hand, parsimonious DCMs are believed to be more interpretable and robust, although their predictive power can be low owing to their misspecification errors. Therefore, it appears to be a natural question whether these two complementary methods can be synergized to retain the strength from both sides. However, since DNNs and DCMs emerged from two different research communities (computer science and economics), it is unclear whether this synergy is even possible, let alone mutually beneficial.

\begin{table}[htb]
\centering
\caption{Comparison of DNNs, DCMs, and TB-ResNets}
\resizebox{1.0\linewidth}{!}{
\begin{tabular}{p{0.55\linewidth}|P{0.15\linewidth}|P{0.15\linewidth}|P{0.15\linewidth}}
\toprule
Models & Prediction & Interpretability & Robustness \\
\midrule
Deep neural networks (DNNs) & High & Low & Low \\
Discrete choice models (DCMs) & Low & High & High \\
Theory-based residual neural networks (TB-ResNets) & High & High & High \\
\bottomrule
\end{tabular}
} %end resizing
\label{table:method_comparison}
\end{table}

To address the aforementioned challenge, this study designs a theory-based residual neural network (TB-ResNet) that synergizes DNNs and DCMs, demonstrating that this synergy is not only feasible but also desirable, leading to a simultaneous improvement in prediction, interpretation, and robustness. This study consists of three main components. We first demonstrate that DNNs align with the RUM framework by briefly recounting McFadden (1974) and Wang et al. (2020) \cite{McFadden1974,WangShenhao2020_asu_dnn}. Second, we present the TB-ResNet framework, which augments DNNs to DCMs to fit the utility residuals with a $(\delta, 1-\delta)$ formulation, resembling the essence of the standard residual network (ResNet) \cite{HeKaiming2016}. This TB-ResNet framework is further elaborated with six interwoven perspectives: architecture design, model ensemble, gradient boosting, regularization, flexible function approximation, and theory diagnosis. The regularization perspective is formally demonstrated by using the state-of-the-art statistical learning theory to illustrate the intuition that DNNs tend to be too complex to capture the reality and DCMs tend to be too simple to do so. Then we design three instances of TB-ResNets using multinomial logit models (MNL-ResNets), prospect theory (PT-ResNets) for risk preference, and hyperbolic discounting (HD-ResNets) for time preference, showing that the simple TB-ResNet framework can incorporate a wide range of DCMs that are part of the utility maximization framework. Lastly, we use empirical testing to determine whether the three instances of TB-ResNets are effective in three datasets, one collected in Singapore and two from Tanaka (2010) \cite{Tanaka2010}. We found that (with some exceptions) the three instances of TB-ResNet can generally improve the overall prediction, interpretability, and robustness of the pure DCMs and DNNs.

The next section reviews related studies. Section 3 introduces the TR-ResNet and its three instances. Section 4 discusses the design of experiments. Section 5 presents the results, and Section 6 concludes and discusses our findings.

\section{Literature Review}
\label{s:2}
\noindent
Individual decision-making has been a classical research question in economics, transportation, marketing, and many other social science and engineering fields. At least three predominant types of DCMs exist: the multinomial logit (MNL) model describing the trade-off between multiple alternatives, prospect theory (PT) models that analyze decision-making under risk and uncertainty, and hyperbolic discounting models that analyze temporal decision-making. McFadden (1974) developed the seminal MNL model based on random utility maximization and applied the model to travel behavior analysis \cite{McFadden1974}. After McFadden (1974), several generations of researchers refined the MNL model by incorporating heterogeneity, endogeneity, and more complicated substitution patterns \cite{Train1980,Train2009,Ben_Akiva1985}. In terms of risk preference, Neumann and Morgenstein \cite{Neumann1944} created the expected utility model to analyze how individuals make decisions with risky inputs. Kahneman and Tversky \cite{Kahneman1979,Kahneman1992} created prospect theory (PT), which addresses abnormalities that cannot be explained by the initial expected utility models \cite{Neumann1944,Pratt1964,Arrow1965,Sydnor2010}. In the last two decades, researchers gradually improved these models by specifying the formulation of reference points or adding more interactions between attributes and probabilities \cite{Tanaka2010,Dhami2016,Koszegi2006}. With regards to time preference, important models include exponential discounting \cite{Samuelson1937}, hyperbolic discounting (HD) \cite{Loewenstein1992}, quasi-hyperbolic discounting \cite{Donoghue1999} and many others. Given the ubiquity of individual decision-making across a massive number of fields, these three types of theories have been widely applied to analyze travel behavior, technology adoption, fuel economy, policy decisions, insurance premiums, procrastination, and self-control \cite{Camerer1989,Nicholson2012,Liu2013,Donoghue2001,Kaur2015}.

Recently researchers have started to use DNNs to predict travel behavior, demonstrating that DNNs can outperform DCMs in terms of prediction accuracy, although these studies often fail to connect DNNs to DCMs in a deeper manner. Individual decision-making can be treated as a ML classification task because the target variables are often discrete. Researchers used DNNs to predict travel mode choice \cite{Cantarella2005}, car ownership \cite{Paredes2017}, travel accidents \cite{ZhangZhenhua2018}, travelers' decision rules \cite{Cranenburgh2019}, driving behaviors \cite{HuangXiuling2018}, trip distribution \cite{Mozolin2000}, hierarchical demand structure \cite{WuXin2018}, queue lengths \cite{LeeSeunghyeon2019}, parking occupancy \cite{YangShuguan2019}, metro passenger flows \cite{HaoSiyu2019}, and traffic flows \cite{Polson2017,LiuLijuan2017,WuYuankai2018,ZhangJunbo2018,DoLoan2019,MaTao2020}. DNNs are also used to complement the smartphone-based survey \cite{XiaoGuangnian2016}, improve survey efficiency \cite{Seo2017}, synthesize new populations \cite{Borysov2019}, and impute survey data \cite{Duan2016}. Studies typically found that the ML classifiers, including DNNs, support vector machines, decision trees, and random forests, can achieve higher predictive performance than the classical DCMs \cite{Pulugurta2013,Omrani2015,Sekhar2016,Hagenauer2017,Cantarella2005}. However, these investigations are mainly limited to a comparative perspective, implicitly intensifying the tension between the data-driven and the theory-driven methods. While several recent studies started to explore the interaction between DNNs and DCMs \cite{WangShenhao2018_stat_learning,WangShenhao2020_asu_dnn,WangShenhao2020_rpsp}, the exploration is still inadequate. Given the predominant use of DCMs and DNNs in travel modeling, it is imperative to demonstrate how to adopt the DNN perspective to analyze individual decision-making beyond prediction. 

Prediction accuracy should not be the only focus, because interpretability and robustness are both important criteria \cite{Lipton2016,Freitas2014,Boshi_Velez2017}. Although many recent studies have focused on DNN interpretation, \cite{Hinton2015,Ribeiro2016,Erhan2009,Baehrens2010,Szegedy2014}, DNNs are still largely perceived as lacking interpretability \cite{Kotsiantis2007}. This is not surprising given that DNNs were initially designed to maximize the prediction power. DNNs and DCMs respectively focus on prediction and interpretation, or equivalently, on predicting $\hat{y}$ and estimating $\hat{\beta}$, as argued by Mullainathan and Spiess (2017) \cite{Mullainathan2017}. In the transportation field, only a small number of studies have touched upon the interpretability issue of DNNs for choice modeling. For example, researchers extracted full economic information from DNNs \cite{WangShenhao2020_econ_info}, ranked the importance of DNN input variables \cite{Hagenauer2017}, or visualized the input-output relationship to improve the understanding of DNN models \cite{Bentz2000}. The challenge of analyzing model interpretability is partially caused by the ambiguity of its definition, as pointed out by Lipton (2016) \cite{Lipton2016}. For example, interpretability can be defined as ``simulatability'': whether researchers can easily simulate the model in their mind. It can also be defined as the capacity to approximate the true probabilistic behavioral mechanism for the choice modeling context \cite{WangShenhao2020_econ_info,WangShenhao2018_stat_learning}. This work adopts the latter definition recognizing the importance of behavioral realism in demand modeling. 

In the choice modeling context, robustness represents the local stability of economic information and the regularity of the behavioral mechanism, which is formally measured by the prediction invariance to random noises or adversarial attacks. When a choice model is robust, a small perturbation of the inputs, such as a \$0.1 decrease in transit fare, does not lead to a dramatic change in outputs, such as a dramatic increase in the choice probability of using public transit. While robustness is not a common topic in the classical demand modeling framework, it is becoming more important for the heavily parameterized DNNs, which are more likely to present irregular local patterns and have been widely criticized as lacking robustness \cite{Szegedy2014,Goodfellow2015,WangShenhao2020_econ_info}. To formally measure robustness, researchers need to examine whether predictive performance significantly decreases under random noises or adversarial attacks. Many adversarial attacks have been created, including the fast gradient sign method (FGSM), one-step target gradient sign method (TGSM), and iterative least-likely class method \cite{Szegedy2014,Goodfellow2015,Papernot2016_2,Papernot2016,Kurakin2016,Kurakin2017}. This study will evaluate the robustness of DCMs, DNNs, and TB-ResNets under both the random noises and the adversarial attacks.

%Szegedy et al. (2014) \cite{Szegedy2014} firstly demonstrated that DNNs do not have local reliability: the adversarial examples, which are perceived the same as initial pictures by human beings, are labeled wrongly by DNNs with high confidence. Researchers can attack DNNs by using a variety of algorithms, including fast gradient sign method (FGSM), one-step target gradient sign method (TGSM), and iterative least-likely class method, as illustrated in \cite{Goodfellow2015,Papernot2016_2,Papernot2016,Kurakin2016,Kurakin2017}. To make the DNNs more robust, researchers also designed defense algorithms, including defensive distillation \cite{Papernot2016}, adversarial training by using both clean and adversarial examples \cite{Kurakin2016}, minimax formulation of robust optimization \cite{Madry2017}, and the input gradient regularizations in training \cite{Ross2018}. 

\section{Theory}
\noindent
Section \ref{sec:s3_dnn_rum} demonstrates that DNNs have an implicit utility interpretation by recounting the results from McFadden (1974) \cite{McFadden1974} and Wang et al. (2020) \cite{WangShenhao2020_asu_dnn}. Subsection \ref{sec:s3_tb_resnet} presents the TB-ResNet framework, and introduces six pertinent ML and behavioral perspectives. Subsection \ref{sec:s3_est_errors} formally illustrates the regularization perspective to illustrate the rationale underlying the design of TB-ResNets. Subsection \ref{sec:s3_three_instances} substantiates this TB-ResNet framework by creating three instances for three choice scenarios.

\subsection{Deep neural networks and random utility maximization}
\label{sec:s3_dnn_rum}
\noindent
The choice analysis includes two types of inputs: alternative-specific variables $x_{ik}$ and individual-specific variables $z_i$ with $i \in \{ 1, 2, ..., N \}$ representing the individual index and $k \in \{ 1, 2, ... K\}$ the alternative index. Let $B = \{ 1, 2, ..., K\}$ and $\tilde x_i =[x^T_{i1}, ..., x^T_{iK}]^T$. The output is individual $i$'s choice, denoted as $y_i = [y_{i1}, y_{i2}, ... y_{iK}]$, with each $y_{ik} \in \{0,1\}$ and $\underset{k}{\sum} y_{ik} = 1$. The RUM framework assumes that individuals maximize the sum of a deterministic utility $v_{ik}$ and a random utility $\epsilon_{ik}$:
\begin{equation}
\setlength{\jot}{2pt} \label{eq:util}
  \begin{aligned}
  u_{ik} = v_{ik} + \epsilon_{ik} = V_{k}(z_i, \tilde x_i) + \epsilon_{ik}
  \end{aligned}
\end{equation}

\noindent in which $v_{ik}$ represents the deterministic utility value and $V_{k}$ the utility function. The probability of individual $i$ choosing alternative $k$ is
\begin{equation}
\setlength{\jot}{2pt} \label{eq:prob}
  \begin{aligned}
  P_{ik} &= Prob(v_{ik} + \epsilon_{ik} > v_{ij} + \epsilon_{ij}, \forall j \in B, \ j \neq k)
  \end{aligned}
\end{equation}

\noindent Assuming $\epsilon_{ik}$ is independent and identically distributed across individuals and alternatives and the cumulative distribution function of $\epsilon_{ik}$ is $F(\epsilon_{ik})$, then

\begin{equation}
\setlength{\jot}{2pt} \label{eq:prob_1}
  \begin{aligned}
  P_{ik} &= \int \underset{j \neq k}{\prod} F_{\epsilon_{ij}}(v_{ik} - v_{ij} + \epsilon_{ik}) d F(\epsilon_{ik})
  \end{aligned}
\end{equation}

\noindent and the two following propositions hold: 

\begin{proposition} \label{prop:choice_prob_from_epsilon_to_softmax} \textit{Suppose $\epsilon_{ik}$ follows a Gumbel distribution, with a probability density function equal to $f(\epsilon_{ik}) = e^{-\epsilon_{ik}} e^{-e^{-\epsilon_{ik}}}$ and a cumulative distribution function equal to $F(\epsilon_{ik}) = e^{-e^{-\epsilon_{ik}}}$. Then the choice probability $P_{ik}$ takes the form of the Softmax activation function.}
\begin{equation}
\setlength{\jot}{2pt} \label{eq:choice_prob}
  \begin{aligned}
  P_{ik} = \frac{e^{v_{ik}}}{\underset{j}{\sum} e^{v_{ij}}}
  \end{aligned}
\end{equation}
\end{proposition}

\begin{proposition} \label{prop:choice_prob_from_softmax_to_epsilon} \textit{Suppose Equation \ref{eq:prob_1} holds, and choice probability $P_{ik}$ takes the form of the Softmax function as in Equation \ref{eq:choice_prob}. If $\epsilon_{ik}$ is a distribution with the transition complete property, then $\epsilon_{ik}$ follows a Gumbel distribution, with $F(\epsilon_{ik}) = e^{-\alpha e^{-\epsilon_{ik}}}$.}
\end{proposition}

Propositions \ref{prop:choice_prob_from_epsilon_to_softmax} and \ref{prop:choice_prob_from_softmax_to_epsilon} jointly demonstrate that DNNs have an implicit RUM interpretation. Specifically, the Softmax activation function, which is used in nearly all DNN architectures as the last layer, implies the random utility terms with a Gumbel distribution under the RUM framework. When a fully connected feedforward DNN is applied to inputs $\tilde x_i$ and $z_i$, the implicit assumption is RUM with the random utility term following Gumbel distribution. Therefore, the inputs into the Softmax activation function in DNNs can be interpreted as the utilities of alternatives. The Softmax function itself is a process of comparing utility scores. The DNN transformation prior to the Softmax function is a process of specifying utilities. Proposition \ref{prop:choice_prob_from_epsilon_to_softmax} can be found in nearly all the textbooks of choice modeling \cite{Train2009,Ben_Akiva1985}, and Proposition \ref{prop:choice_prob_from_softmax_to_epsilon} is from Lemma 2 in \citeauthor{McFadden1974} (\citeyear{McFadden1974}). By taking advantage of the RUM interpretation in DNNs, researchers can design novel DNN architectures to improve the model performance \cite{WangShenhao2020_asu_dnn}. The sketch proof of the two propositions is in Appendix I. 

DNNs and DCMs share the utility maximization framework, but they parameterize their utility functions in different ways. DNNs can automatically learn utility function owing to the strong approximation power of complex DNNs' model families \cite{Hornik1989,Hornik1991,Cybenko1989}, while DCMs rely on much more parsimonious parametric assumptions. For example, the utility function of DNNs ($V_{DNN,k}$) can be parameterized by millions of parameters, while that of DCMs ($V_{T,i}$) often by less than ten parameters. The similar utility interpretation shared by DNNs and DCMs enables us to design the TB-ResNet framework, and their differences in model complexity are an opportunity for complementarity. 

\subsection{Theory-based residual neural networks}
\label{sec:s3_tb_resnet}
\noindent
Leveraging the similar utility interpretation in DCMs and DNNs, we design the TB-ResNet framework, which consists of a DCM utility function $V_{T,k}(z_i, \tilde x_i)$ and a DNN utility function $V_{DNN,k}(z_i, \tilde x_i)$ weighted by $\delta$ and 1-$\delta$:
\begin{flalign}
v_{TB-ResNet,ik} = (1-\delta) v_{T,ik} + \delta v_{DNN, ik} = (1-\delta) V_{T,k}(z_i, \tilde x_i) + \delta V_{DNN, k}(z_i, \tilde x_i)
\label{eq:tb_resnet_util}
\end{flalign}

\begin{figure}[ht]
\centering
\includegraphics[width=0.6\linewidth]{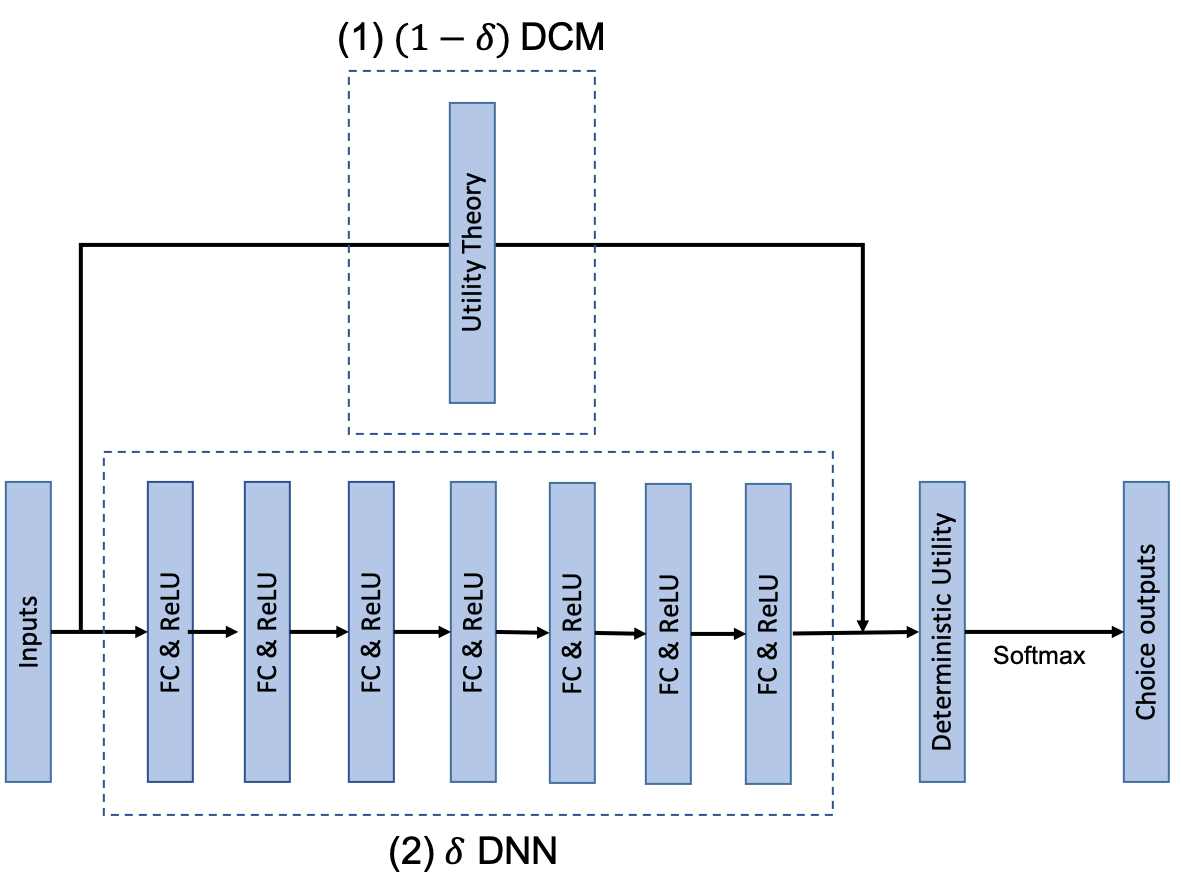}
\caption{Architecture of TB-ResNet. Both DCM and DNN are flexible: the DNN block uses seven layers as an example, but it can be any depth or width; the DCM block can take any utility specification under the RUM framework.}
\label{fig:arch}
\end{figure}

\noindent
where $V_{T,k}$ represents the utility function from DCMs, $V_{DNN, k}$ represents that from DNNs, and ($\delta$, 1-$\delta$) adjusts the weighting between them. This TB-ResNet can be seen as a linear combination of the two types of utility functions with a flexible weighting controlled by $\delta$. The utility specification of a feedforward DNN $V_{DNN, ik}$ in Equation \ref{eq:tb_resnet_util} can be parameterized as 
\begin{equation}
\setlength{\jot}{2pt} \label{eq:util_specification}
  \begin{aligned}
  V_{DNN,k}(z_i, \tilde x_i) = w_{m,k}^T \Phi(z_i, \tilde x_i) = w_{m,k}^T (g_{m-1} ... \circ g_2 \circ g_1)(z_i, \tilde x_i) \\
  \end{aligned}
\end{equation}
\noindent in which $m$ is the number of layers of DNN, $g_l(t) = ReLU(W_l^T t + b_l)$, and $ReLU(t) = \max(0, t)$. The $W_l$ represents the parameters in DNNs for layer $l$, and the $ReLU$ is one activation function among many alternatives (e.g. Tanh and Sigmoid). The DCM utility specification can be parameterized by various utility theories, which we will discuss in Section \ref{sec:s3_three_instances}. 

Figure \ref{fig:arch} represents the architecture of the TB-ResNet, consisting of the shallow $(1-\delta)$DCM and the deep $\delta$DNN blocks for joint specification of the deterministic utility term. Typically the DCM block $(1-\delta)V_{T,k}$ can be represented by a shallow neural network with one-layer transformation, while the DNN block $\delta V_{DNN,k}$ is represented by a deep structure capable of automatic learning. The DCM and DNN blocks transform the inputs ($z_i, \tilde x_i$) into deterministic utilities, which are further converted into choice probabilities and outputs through the Softmax activation function. This TB-ResNet framework can be understood from six interwoven ML and behavioral perspectives as follows.

First and most intuitively, this TB-ResNet can be seen as a new DNN architecture, because the DCM part in the TB-ResNet represents a shallow neural network and the DNN part represents a deep one. In fact, the name of the TB-ResNet arises from the standard ResNet architecture, which consists of an identity feature mapping and a feedforward DNN architecture: $v_{ResNet,ik} = V_{I,k}(z_i, \tilde x_i) + V_{DNN, k}(z_i, \tilde x_i)$, where $V_{I,k}(x) = x$. When the true model is close to linear, the ResNet can approximate the true model better than a standard feedforward DNN. This reasoning can be similarly applied to TB-ResNets. When $1-\delta$ is close to one, the TB-ResNet consists of a main DCM part and a small DNN part fitting the utility residual, resembling the essence of the standard ResNet architecture.

Second, the TB-ResNet framework with the $(\delta, 1-\delta)$ weighting can be seen as an ensemble model of DCMs and DNNs with scale adjustment. The weighting is controlled by the ratio $\frac{1-\delta}{\delta}$, which can span all possible positive values under a logarithmic scale of $\delta \in (0, 1)$.\footnote{The logarithmic scale refers to the $\delta$ taking the values of $10^{-x}$ and $1-10^{-x}$ as it gets close to zero or one. This design is intended to maximize the span of the magnitude of the scale ratio $\frac{1-\delta}{\delta}$.} When $\delta \rightarrow 0$, such as $\delta = 10^{-5}$, the utility ratio $\frac{1-\delta}{\delta}$ converges to $+\infty$; when $\delta \rightarrow 1$, such as $\delta = 1-10^{-5}$, the utility ratio $\frac{1-\delta}{\delta}$ converges to $0$; when $\delta = 0.5$, the ratio equals to one. Therefore, this $(\delta, 1-\delta)$ weighting allows us to explore all the possible utility ratio of the DCM and the DNNs. In fact, the flexible scaling is closely related to the randomness discussion in classical choice modeling. For example, to combine revealed and stated preference data, researchers need to adjust a scale factor to reflect the different randomness in two types of data sets \cite{WangShenhao2020_rpsp,Hensher1993,Bradley1997}. 

Third, the TB-ResNet framework is similar to the gradient boosting method \cite{Friedman2001}, although differences still exist. They are similar since both seek to achieve higher performance by adding multiple models; in particular, the TB-ResNets with a sequential training procedure are similar to the boosting method with multiple stages. However, they are also different since gradient boosting is typically used to combine multiple weak classifiers, while TB-ResNets combine a relatively weak classifier (DCMs) and a strong one (DNNs). As a result, the regularization perspective becomes essential in the TB-ResNets, particularly when $\delta$ is small. In addition, the TB-ResNets combine the DCMs and DNNs through the shared utility interpretation, while the boosting method connects multiple classifiers through the multi-stage optimization of the loss function. The shared utility interpretation in TB-ResNets is helpful in obtaining not only lower prediction losses, but also improvements in local regularity, robustness, and utility-based economic interpretation. Nonetheless, the authors acknowledge that it a fine line to differentiate between model ensemble, gradient boosting, and our TB-ResNet framework. It is also possible to improve the TB-ResNet framework by incorporating the other two perspectives in the future. 

Fourth, when $\delta \rightarrow 0$, the TB-ResNet is dominated by the DCM component, which becomes a skeleton utility function to localize and stabilize the TB-ResNet system, and the small $\delta$ regularizes the complex DNN component to address overfitting. With a small $\delta$, the model complexity between $(1-\delta) V_{T,k}$ and $\delta V_{DNN,k}$ is more balanced, since the DNN component is typically much more complex than the DCM component. Intuitively, when $\delta$ becomes smaller, the TB-ResNet framework is increasingly localized around the DCM component, and the training of the DNN is similar to a searching around the small neighbor of the DCM. A small $\delta$ is most effective when simple DCMs can successfully capture much of the true behavioral mechanism. In fact, when the DCM component can perfectly capture the true behavioral mechanism, the best $\delta$ should be close to zero.

Fifth, when $\delta \rightarrow 1$, the TB-ResNet is dominated by the DNN component, which allows the TB-ResNet system to use the outstanding approximation power of DNNs to approximate the true data generating process, transcending the theoretical limitations of the DCMs. When $\delta$ is closer to one, it trends towards a large regional search by the $\delta$DNN component around the small $(1-\delta)$DCM component. A large $\delta$ is the most effective when simple DCMs capture little information about the decision-making mechanism. In the worst scenario,  when the DCMs capture zero information, the TB-ResNet reduces to a DNN model. 

Therefore, the optimum $\delta$ value becomes a metric to diagnose the completeness of the DCMs. A small and optimum $\delta$ suggests that the current DCM is highly effective since only a small portion of the DNN component is needed to fit the utility residuals. A large and optimum $\delta$ suggests that the current DCM is far from complete since the TB-ResNet mainly uses the DNN component to fit the true behavior. Therefore, the optimum $\delta$ value can be a tool to diagnose the completeness of the DCMs. However, the optimum $\delta$ value can only be identified empirically since modelers cannot evaluate the completeness of a theory a priori. The results section will compare the optimum $\delta$ values of our three instances, which sheds light on the theoretical completeness of the MNL, PT, and HD models. 

In summary, simple DCMs tend to underfit the true behavioral mechanism, while rich DNNs tend to overfit. The TB-ResNets are formulated with a flexible $(\delta, 1-\delta)$ weighting, taking advantage of both the simplicity of DCMs and the richness of DNNs, and guarding against problems from both sides. A large $\delta$ enables the DNN component to fit the utility residual of the DCM component to address the underfitting problem, and a small $\delta$ controls the scale of the DNN component as a regularization tool to address the overfitting problem. This is the key intuition underlying the design of TB-ResNets.

%By its design, it is expected that this TB-ResNet can improve the prediction accuracy, interpretability, and robustnes of both DNNs and DCMs. Compared to a pure theory-driven model, the insights from theory is maintained in the $V_{T,ik}$ of the TB-ResNet, and the second-stage training of TB-ResNet can improve the prediction accuracy since the DNN part of the TB-ResNet can address the misspecification error that commonly exists in any theory-based model. The second-stage of TB-ResNet can also improve the interpretability by augmenting the $\delta V_{DNN, ik}$ utility function to $V_{T,ik}$, rendering the final utility function richer than the simple $V_{T,ik}$. Compared to a DNN model that does not have regular local information, TB-ResNet is more interpretable owing to the theory-based utility function, which stablizes the local information for the whole input domain. Therefore, TB-ResNets should be more robust than DNNs regarding the adversarial examples, since it would be more difficult to create adversarial examples in the local region of each instance for the TB-ResNets when the local information is stablized by the theory-based utility function in TB-ResNets. TB-ResNets are also likely to improve the prediction accuracy of DNNs, when the decision-making theory is informative in capturing the true behavior. In short, as TB-ResNets combine decision-making theories and DNNs, we expect TB-ResNet to incorporate the strength from both sides and address the weaknesses for each other.

\subsection{Balancing approximation and estimation errors of DCMs and DNNs}
\label{sec:s3_est_errors}
\noindent
Formally, the problems of underfitting and overfitting can be framed as the challenge of balancing approximation and estimation errors. This subsection will use the state-of-the-art statistical learning theory to illustrate the importance of the $\delta$ term in balancing the model complexities between DCMs and DNNs. 

The out-of-sample performance of TB-ResNets can be decomposed into approximation and estimation errors, and the analysis into the latter illustrates the importance of controlling model complexity. Let ${\F}_1$ and ${\F}_2$ denote the model families of DCMs and DNNs. The empirical risk minimization (ERM) is used for model training: 
\begin{flalign} \label{eq:erm}
\hat{f} = \argmin_{f \in (1-\delta) {\F}_1 + \delta {\F}_2} \frac{1}{N} \sum_{i = 1}^N l(y_i, f(x_i))
\end{flalign}

\noindent in which $x_i$ is a vector representing both the alternative-specific inputs $x_{ik}$ and the individual-specific inputs $z_i$. Let $f^*$ denote the true data generating process. The \textit{Excess error} is defined and decomposed as following:
\begin{flalign}
{\E}_S [L(\hat f) - L(f^*)] &= {\E}_S[L(\hat f) - L(f^*_{\F})] + {\E}_S[L(f^*_{\F}) - L(f^*)]
\label{eq:excess_error_decomposition}
\end{flalign}

\noindent
where $L = {\E}_{x,y} [l(y, f(x)]$ is the expected loss function and $S$ represents the sample $\{x_i, y_i \}_1^N$; $f^*_{\F} = \underset{f \in \F }{\argmin} \ L(f)$, the best function in function class $\F := (1-\delta){\F}_1 + \delta{\F}_2$ to approximate $f^*$. The excess error measures the average difference of the out-of-sample performance between the estimated function $\hat{f}$ and the true model $f^*$. The excess error is decomposed as an \textit{estimation error} ${\E}_S[L(\hat f) - L(f^*_{\F})]$ and an \textit{approximation error} ${\E}_S[L(f^*_{\F}) - L(f^*)]$. The approximation error is deterministic and is thus irrelevant to the training procedure, so this study does not address approximation errors in details. A simple model family (e.g. DCMs) will usually approximate the true data generating process worse than a complex model family (e.g. DNNs), which can also be inferred from the universal approximator theorem of the DNN \cite{Hornik1989}. The key upper bound is on the estimation error ${\E}_S[L(\hat f) - L(f^*_F)]$, which is provided by using Rademacher complexity from statistical learning theory.

\begin{definition} \label{def:1}
Empirical Rademacher complexity of function class $\F$ is defined as:
\begin{flalign}
\hat{\R}_n({\F}|_S) = {\E}_{\epsilon} \ \underset{f \in {\F}}{\sup} \ \frac{1}{N} \sum_{i=1}^N \epsilon_i f(x_i) 
\end{flalign}
$\epsilon_i$ is the Rademacher random variable, taking values $\{-1, +1\}$ with equal probabilities.
\end{definition}

\begin{proposition} \label{prop:3}
\textit{The estimation error of an estimator $\hat{f}$ can be bounded by the Rademacher complexity of $\F$.}
\begin{flalign}
{\E}_S[L(\hat f) - L(f^*_{\F})] \leq 2{\E}_S \hat{\R}_n({\F}|_S) 
\end{flalign}
\end{proposition}

Definition \ref{def:1} and Proposition \ref{prop:3} jointly provide the intuition that the upper bound of the estimation error of any estimate $\hat{f}$ can be approximated by the complexity of the model family $\F$. Note that $\hat{\R}_n({\F}|_S)$ in Definition \ref{def:1} measures how complex the model family $\F$ is, and the averaged Rademacher complexity on the right side of Proposition \ref{prop:3} is the upper bound on the estimation error. In other words, it is important to limit the complexity of the model family $\F$ to achieve a tight upper bound on the estimation error. The proof of Proposition \ref{prop:3} can be found in \cite{WangShenhao2018_stat_learning}, and the key technique is the symmetrization lemma \cite{Wainwright2019}.

In the case of TB-ResNets, $\F$ is designed to be $(1-\delta){\F}_1 + \delta {\F}_2$. The following three propositions provide the intuition as to why the $\delta$ weighting is important:
\begin{proposition} \label{prop:4}
\textit{The estimation error of the TB-ResNet estimator $\hat{f}$ can be bounded by the weighted Rademacher complexities of ${\F}_1$ and ${\F}_2$.}
\begin{flalign}
{\E}_S[L(\hat f) - L(f^*_{\F})] \leq 2{\E}_S [(1-\delta)\hat{\R}_n({\F}_1|_S) + \delta \hat{\R}_n({\F}_2|_S)]
\end{flalign}
\end{proposition}

\begin{proposition} \label{prop:5}
The Rademacher complexity of the DCM model family ${\F}_1$ can be bounded by
\begin{flalign}
\hat{\R}_n({\F}_1|_S) \lesssim O(\sqrt{\frac{v}{N}})
\end{flalign}
in which $v$ is the VC dimension of function class ${\F}_1$ and $N$ is the sample size.
\end{proposition}

\begin{proposition} \label{prop:6}
The Rademacher complexity of the DNN model family ${\F}_2$ can be bounded by 
\begin{flalign}
\hat{\R}_n({\F}_2|_S) \lesssim \frac{(\sqrt{2 \log(D)} + 1) \sqrt{\frac{1}{N} \sum_{i=1}^N ||x_i||^2}}{\sqrt{N}} \times \prod_{j = 1}^D M_F(j)
\end{flalign}
in which $D$ represents the depth of DNN, $||x_i||$ represents the norm of input variables, and $M_F(j)$ is the upper bound of the Frobenius norm of parameter matrix $W_j$. Here the DNN model uses ReLU activation functions.
\end{proposition}

Propositions \ref{prop:4}, \ref{prop:5}, and \ref{prop:6} demonstrate the importance of using $\delta$ to balance the estimation errors between ${\F}_1$ and ${\F}_2$. Since the DNN model family ${\F}_2$ is much more complicated than the DCM model family ${\F}_1$, a strong regularization needs to be imposed on the DNN model family ${\F}_2$ to guarantee a balanced complexity between $(1-\delta)\hat{\R}_n({\F}_1|_S)$ and $\delta \hat{\R}_n({\F}_2|_S)$. Specifically, the DNN complexity, as shown in Proposition \ref{prop:6}, is exponential in the depth $D$ and also depends on the size of the input variables $x_i$. On the other hand, the DCM model ${\F}_1$ only depends on the VC dimension, which is often linearly related to the number of parameters. Therefore, the DNN complexity is typically much larger than the DCM complexity ($\hat{\R}_n({\F}_2|_S) >> \hat{\R}_n({\F}_1|_S)$). In this case, a small $\delta$ weighting on the DNN part can limit the total complexity of the TB-ResNets ($(1-\delta){\F}_1 + \delta{\F}_2$), thus improving the model performance. The proof of Proposition 5 can be found in Wang et al. (2020) \cite{WangShenhao2020_econ_info}, and that of Proposition 6 is in Golowich et al. (2017) \cite{Golowich2017}. We provide a sketch proof for Proposition \ref{prop:4} in Appendix IV. 

\subsection{Three instances of TB-ResNets}
\label{sec:s3_three_instances}
\subsubsection{Multinomial logit based residual neural networks (MNL-ResNets)}
\noindent
The TB-ResNet framework is flexible because the DCM utility $V_{T,{k}}$ can take different forms depending on the context. In the MNL setting, we use the linear utility specification as the theory part of TB-ResNets. Replacing the $V_{T,k}$ in Equation \ref{eq:tb_resnet_util} by $V_{MNL,k}$ and removing index $i$ for simplicity, the MNL-ResNet is formulated as 
\begin{flalign}
& v_{MNL-ResNet,k} = (1-\delta)v_{MNL,k} + \delta v_{DNN, k} = (1-\delta)V_{MNL,k}(z, \tilde x) + \delta V_{DNN, k}(z, \tilde x) \label{eq:pt_util} \\
& V_{MNL,k}(z, \tilde x) = w_{0,k} + w_{x_k}' x_k + w_z' z 
\end{flalign}

\noindent
where $w_{x_k}$ represents the parameters for the alternative-specific variables and $w_z$ represents the parameters of the individual-specific variables. This linear-in-parameter specification of $V_{MNL, k}$ is widely used in choice modeling. 

\subsubsection{Prospect theory based residual neural networks (PT-ResNets)}
\noindent
In the risk preference setting, we use prospect theory as the theory part of TB-ResNets. Replacing $V_{T,k}$ in Equation \ref{eq:tb_resnet_util} by $V_{PT,k}$, the utility specification of PT-ResNet is formulated as 
\begin{flalign}
& v_{PT-ResNet,k} = (1-\delta)v_{PT,k} + \delta v_{DNN, k} = (1-\delta)V_{PT,k}(z, \tilde x) + \delta V_{DNN, k}(z, \tilde x) \label{eq:pt_util} \\
& V_{PT,k}(z, \tilde x) = \sum_j c(x_{kj})\pi(p_{kj})
\end{flalign}
Then the value function $c(x_{kj})$ and probability weighting function $\pi(p_{kj})$ are further parameterized as
\begin{flalign}
& c(x_{kj}) = 
\Big\{
  \begin{tabular}{ccc}
  $x_{kj}^r$ & $x_{kj} \geq 0$ \\
  $-\lambda (-x_{kj})^r   $ & $x_{kj} \leq 0$ \\
  \end{tabular} \\
& \pi(p_{kj}) = e^{-(- ln \ p_{kj})^\alpha} \\
& \alpha = \alpha(z) = \alpha_0 +  z' w_{\alpha z} \\
& r = r(z) = r_0 + z' w_{r z} \\
& \lambda = \lambda(z) = \lambda_0 + z' w_{\lambda z} \label{eq:pt_parameter_lambda}
\end{flalign}

\noindent
In the equations above, $j$ is the index of uncertain monetary payoffs; $x_{kj}$ is the monetary payoff for alternative $k$ at the value indexed by $j$; $p_{kj}$ is the winning probability of $x_{kj}$; $\alpha$ represents the probability weighting factor; $r$ represents the concavity of value functions; $\lambda$ represents the loss aversion factor; $\alpha$, $r$, and $\lambda$ are individual specific and can be partially explained by socioeconomic variables $z$. The specification in Equations from \ref{eq:pt_util} to \ref{eq:pt_parameter_lambda} is basically the same as Tanaka et al. (2010) \cite{Tanaka2010}.\footnote{There are two slight differences between our PT model and Tanaka et al. (2010): the initial paper used a non-parametric method to estimate individuals' risk preference parameters and sequentially estimated the coefficients $w_{\alpha z}$, $w_{r z}$, and $w_{\lambda z}$, while our PT model follows a parametric method and simultaneously estimates all the coefficients.} PT is widely used for travel behavioral analysis because travel decisions often involve time uncertainty \cite{Palma2008,Avineri2008}. 

%Besides, Equations from \ref{eq:pt_util} to \ref{eq:pt_parameter_lambda} actually reveal deeper relationship between utility theory and DNN. PT model contains two types of information. The first type could be called \textit{meta-architecture} information, referring to the layer-by-layer feature mapping from $x, p, z$, to values $v$ and decision weights $\pi$, and to output layer utility $V_{PT,k}$. The second type is the parametric specification of constituent functions, such as the power form of value function. This two types of information are similar to DNN, which also consists of global meta-architecture and local constituent functions.

\subsubsection{Hyperbolic discounting based residual neural networks (HD-ResNets)}
\noindent
Another instance of TB-ResNets is the hyperbolic discounting residual network (HD-ResNet) for time preferences. The utility function of the HD-ResNet is formulated as following:
\begin{flalign}
& v_{HD-ResNet,k} =(1-\delta) v_{HD,k} + \delta v_{DNN, k} = (1-\delta)V_{HD,k}(z, \tilde x) + \delta V_{DNN, k}(z, \tilde x) \label{eq:hd_util} \\
& V_{HD,k}(z, \tilde x) = \sum_j x_{kj} \beta e^{-rt_{kj}} 
\end{flalign}
\noindent Then $\beta$ and $r$ can be further parameterized as 
\begin{flalign}
& \beta = \beta(z) = \beta_0 +  z' w_{\beta z} \label{eq:hd_parameter_beta} \\
& r = r(z) = r_0 + z' w_{r z} \label{eq:hd_parameter_r}
\end{flalign}
\noindent
In the equations above, $x_{kj}$ is the monetary payoff; $t_{kj}$ is the associated time; $r$ is the conventional time discounting factor; $\beta$ is the present-bias factor; both $r$ and $\beta$ can be partially explained by socioeconomic variables $z$. The specifications from Equation \ref{eq:hd_util} to \ref{eq:hd_parameter_r} are the same as Tanaka et al. (2010) \cite{Tanaka2010}. 

%As a summary, researchers can use three different methods to approach decision-making questions. The first is a pure theory-driven method (PT and HD); the second is a pure data-driven method (DNN); and the third is the TB-ResNet that combines decision-making theory and DNN with a sequential training procedure. This synthesis is feasible due to Propositions \ref{prop:choice_prob_from_epsilon_to_softmax} and \ref{prop:choice_prob_from_epsilon_to_softmax}, and this synthesis is potentially beneficial since the TB-ResNet maintains the interpretability of the theory and the prediction power of DNN. To the best of our knowledge, this study is the first one that introduces the perspective of synthesizing interpretable and prediction models in the ResNet framework. This perspective is quite generic, since any interpretable utility theory could be combined with the prediction-driven DNN, as long as the utility theory is within the framework of utility maximization.

\section{Experiment Setup}
\subsection{Datasets}
\noindent The experiments use three datasets for the three instances. The first data set (SG) is a stated preference survey collected by the authors in Singapore in 2017, focusing on the travel mode choice between walking, buses, ridesharing, driving, and autonomous vehicles. The second and third datasets are stated preference surveys collected from Tanaka et al. (2010) \cite{Tanaka2010}, which focused on the risk and time preference of two monetary alternatives. 

The summary statistics of the three data sets are provided in Appendix II. The sample sizes of the three datasets are respectively $8,418$, $6,335$ and $5,340$, which are similar to typical travel behavioral surveys. The choices in the three data sets are all balanced: even in the most unbalanced case, the smallest share is the $10.38\%$ of the respondents who chose to walk in Singapore. The attributes in the SG data set were designed by using the standard orthogonal experiment design based on the average travel time and costs of the travel alternatives in Singapore. The attributes in the PT and HD data sets were designed by using simulations based on PT and HD theories. The survey for the SG data set was conducted with the help of an online company Qualtrics; the surveys for the PT and HD data sets were collected through interviews with the help of local officials. For data collection details of the three data sets, readers can refer to \cite{WangShenhao2019_risk_av} and \cite{Tanaka2010}. %Readers could find the summary statistics of socioeconomic variables in Tanaka et al. (2010), and the results of our PT and HD models in Appendix II. In short, our parameters estimated from PT and HD are close to Tanaka et al. (2010). 

\subsection{Training}
\noindent 
The three data sets are divided into training and testing sets, with the former used to train the TB-ResNets and the latter to demonstrate the effects of different $\delta$ values. The sample sizes of the training and testing sets in the SG, PT, and HD data sets are (5050, 1648), (4199,1050), and (4272, 1068), respectively. To concentrate the focus on the $\delta$ factor, we largely simplified the hyper-parameter searching of the DNN component by using only two simple DNN architectures. The first DNN architecture is designed with depth = 3, width = 100, number of iterations = 5,000, and size of mini-batch = 100, and the second DNN architecture is the same as the first except with depth = 5. The two DNN architectures are compared to reveal that the TB-ResNets performance partially depends on the configuration of the DNN part. However, this work mainly seeks to demonstrate the benefits through the synergistic TB-ResNet framework rather than the improvements \textit{within} either the DCM or the DNN component. It is because either the DCM or the DNN component can be improved by a vast number of methods, given the past four decades' work on the DCMs and the recent popularity of the DNNs, which are beyond the scope of this work. Our empirical result seeks to demonstrate that, conditioning on any given DCM or DNN, the TB-ResNets can generate mutual benefits for both components. 

The effects of $\delta$ values are demonstrated in the testing sets and explored on a logarithmic scale.\footnote{The $\delta$ spans a list of values: [1e-10,1e-8, 1e-7,1e-6,1e-5, 1e-4, 0.001, 0.002, 0.004, 0.005, 0.006,0.007, 0.008,0.009, 0.01,0.03, 0.05, 0.1, 0.3, 0.5, 0.8, 0.9,0.95,0.99,0.999,0.9999, 1]} A small $\delta$ such as $10^{-5}$ postulates a large utility ratio ($\frac{1-\delta}{\delta}$) between the DCM and the DNN parts as $10^5$, while a large $\delta$ such as $1-10^{-5}$ postulates a small ratio as $10^{-5}$. The range of $\delta$ values can assist in empirically evaluating the completeness of the DCM theory, which is impossible to know a priori.\footnote{For application, the authors suggest dividing the datasets into training, validation, and testing sets. The validation set can be used to choose the optimum $\delta$ value and the testing set for application. Since this work focuses on demonstrating the effects of $\delta$, we decided to simplify the data processing into the division of training and testing sets.} An empirically optimum and small $\delta$ suggests the relative completeness of the DCM theory, while an optimum and large $\delta$ suggests that the DCM model is far from complete and needs further development. The optimality of the $\delta$ values will be discussed in all three instances of the TB-ResNets in our result section. 

The three instances of TB-ResNets can be trained in a simultaneous or sequential manner. The simultaneous training implies the the training of ${V}_{DNN, k}$ and $V_{T,k}$ at the same time with the stochastic gradient descent. The sequential training implies a sequential approach of training $V_{T,k}$ on the first stage and training ${V}_{DNN,k}$ on the second. While both methods are consistent with our ML and behavioral perspectives in the TB-ResNet framework, we tend to believe that sequential training is more intuitive than simultaneous training. This is because the DNN component in the TB-ResNet might easily capture all the valuable information that could have been explained by $V_{T,k}$, when $V_{T,k}$ and $V_{DNN,k}$ are simultaneously trained. In this case, the simultaneous training might damage the capacity of $V_{T,k}$ to stabilize the utility function. Nonetheless, we provide the results of both sequential and simultaneous training in our results section and appendices. 

By using $w_T$ to represent the parameters in the DCM part and $w_{DNN}$ to represent the parameters in the DNN part, the formulation of the sequential training procedure is described as follows. In the first stage, empirical risk minimization (ERM) is formulated as shown:
\begin{flalign}
\label{eq:erm_1}
& \underset{V_{T,k} \in {\F}_1}{\min} \ L(\tilde{x}_i, z_i; w_T) = \underset{V_{T,k} \in {\F}_1}{\min} - \frac{1}{N} \sum_{i=1}^{N} \sum_{k=1}^{K} y_{ik} \log \ \frac{e^{(1-\delta)V_{T,k}(\tilde{x}_i, z_i; w_T)}}{{\sum_{j=1}^K} e^{(1-\delta)V_{T,j}(\tilde{x}_i, z_i; w_T)}}
\end{flalign}

\noindent which is the same as the maximum likelihood estimation used in classical choice models. The second stage is another training conditioned on $\hat{w}_{T}$. With ${\F}_2$ representing the model family of DNNs, second-stage training is:
\begin{flalign}
\label{eq:erm_2}
  & \underset{V_{DNN,k} \in {\F}_2}{\min} \ L(\hat{w}_{T}, \tilde{x}_i, z_i; w_{DNN}) = \\ & \ \ \ \ \ \ \underset{V_{DNN,k} \in {\F}_2}{\min}  - \frac{1}{N} \sum_{i=1}^{N} \sum_{k=1}^{K} y_{ik} \log \ \frac{e^{(1-\delta)\hat{V}_{T,k}(\tilde{x}_i, z_i)+ \delta V_{DNN,k}(\tilde{x}_i, z_i; w_{DNN})}}{\sum_{j=1}^{K} e^{(1-\delta)\hat{V}_{T,j}(\tilde{x}_i, z_i) + \delta V_{DNN,j}(\tilde{x}_i, z_i; w_{DNN})}}
\end{flalign}

To evaluate predictive performance, this work uses three metrics: prediction accuracy, cross-entropy loss, and F1 score. Prediction accuracy is formulated as:
\begin{flalign}
Accuracy = 1 - \frac{1}{N} \sum_{i=1}^{N} \I\{ \hat{y}_{i} \neq y_{i} \}
\end{flalign}
\noindent in which $\hat{y}_{i}$ represents the vector of the predicted choice, $\I\{\}$ is an indicator function, and the inequality sign implies that the two vectors ($\hat{y}_{i}$ and $y_{i}$) are different. The cross-entropy loss is formulated as:
\begin{flalign}
Cross \ entropy \ loss = - \frac{1}{N} \sum_{i=1}^N \sum_{k=1}^K y_{ik} \log \hat{P}_{ik}
\end{flalign}
\noindent in which $\hat{P}_{ik}$ is the predicted choice probability from DCMs, DNNs, or TB-ResNets. This cross-entropy loss is also named as log-loss, the same as the negative value of the log likelihood. Lastly, the F1 score is:
\begin{flalign}
F1 \ score = \sum_{k =1}^{K} \mathbbm{W}_k \cdot 2 \times \frac{Precision_k \times Recall_k}{Precision_k + Recall_k}
\end{flalign}
where $\mathbbm{W}_k$ is the share of label k in the sample ($\mathbbm{W}_k =\frac{1}{N}\sum_{i=1}^N y_{ik}$). In the equation above, $Precision_k$ and $Recall_k$ are class-specific and formulated as:
\begin{flalign}
Precision_k = \frac{\sum_{i=1}^N\I\{ \hat{y}_{ik} = y_{ik} \}}{\sum_{i=1}^N  \hat{y}_{ik}  } 
\end{flalign}
\begin{flalign}
Recall_k = \frac{\sum_{i=1}^N\I\{ \hat{y}_{ik} = y_{ik} \}}{\sum_{i=1}^N  {y}_{ik}  } 
\end{flalign}
\noindent Intuitively, $Precision_k$ and $Recall_k$ measure the column- and row-specific performance of the confusion matrix. The F1 score combines the column- and row-specific perspectives with a class-specific weighting. Our three metrics, including accuracy, cross-entropy loss, and F1 score, have taken into account the deterministic and probabilistic decision rules for both balanced and unbalanced outputs. 

\section{Results}
\label{sec:results}
\noindent
The result section starts with visualizion of the utility functions of DCMs, DNNs, and TB-ResNets, thus providing intuition into why DCMs and DNNs are complementary and how TB-ResNets strike a middle ground that retains benefits from both sides. Then we will successively delve into interpretability, prediction, and robustness for comparison and evaluation. 

\subsection{Utility functions of TB-ResNets as combination of DCMs and DNNs}
\label{sec:util_combination}
\begin{figure}[H]
\captionsetup[subfigure]{justification=centering}
\centering
\subfloat[MNL $(50.6\%)$]{\includegraphics[width=0.2\linewidth]{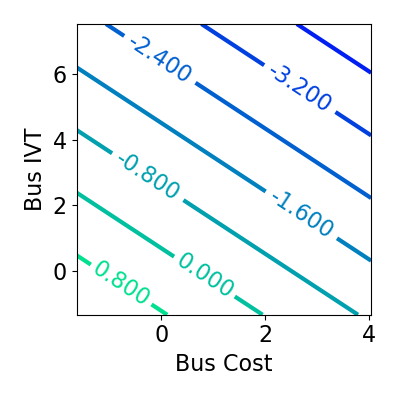}\label{sfig:cm_est_field}} 
\subfloat[MNL-ResNet ($\delta = 10^{-5}$; $53.1\%$)]{\includegraphics[width=0.2\linewidth]{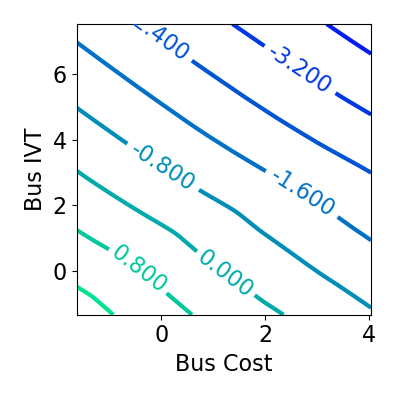}\label{sfig:cm_resnet_1e-5_field}}
\subfloat[MNL-ResNet ($\delta = 0.008$; $57.0\%$)]{\includegraphics[width=0.2\linewidth]{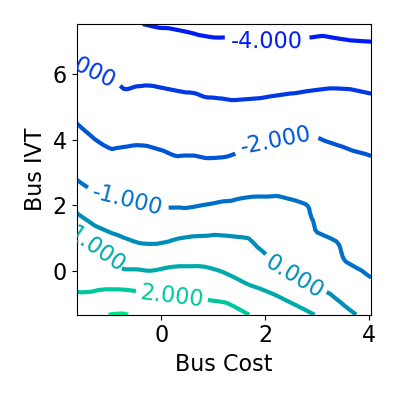}\label{sfig:cm_resnet_0008_field}}
\subfloat[MNL-ResNet ($\delta = 0.05$; $56.1\%$)]{\includegraphics[width=0.2\linewidth]{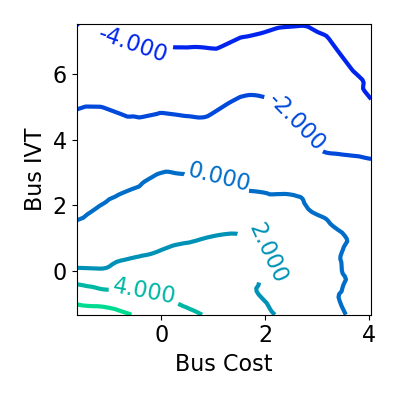}\label{sfig:cm_resnet_001_field}}
\subfloat[DNN ($55.8\%$)]{\includegraphics[width=0.2\linewidth]{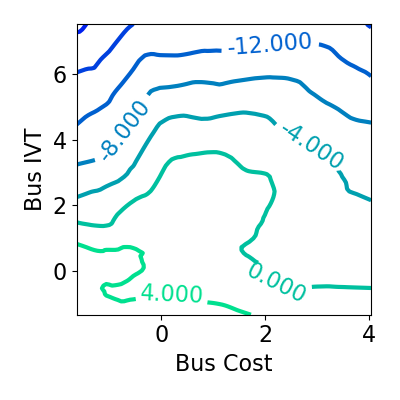}\label{sfig:cm_dnn_field}} \\
\subfloat[Cost]{\includegraphics[width=0.1\linewidth]{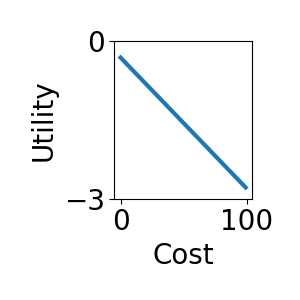}\label{sfig:cm_est_x0}} 
\subfloat[IVT]{\includegraphics[width=0.1\linewidth]{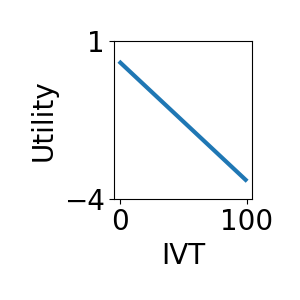}\label{sfig:cm_est_x1}}
\subfloat[Cost]{\includegraphics[width=0.1\linewidth]{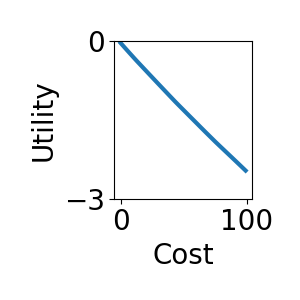}\label{sfig:cm_resnet_1e-5_x0}}
\subfloat[IVT]{\includegraphics[width=0.1\linewidth]{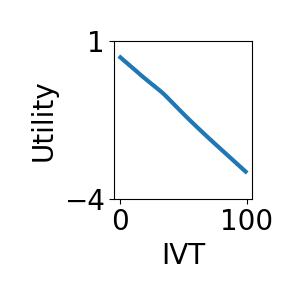}\label{sfig:cm_resnet_1e-5_x1}} 
\subfloat[Cost]{\includegraphics[width=0.1\linewidth]{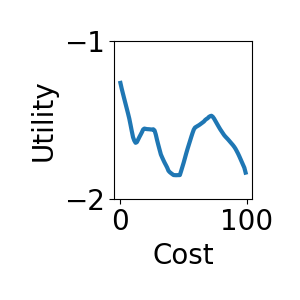}\label{sfig:cm_resnet_0005_x0}}
\subfloat[IVT]{\includegraphics[width=0.1\linewidth]{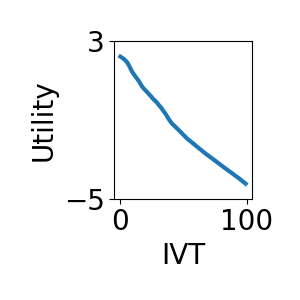}\label{sfig:cm_resnet_0005_x1}}
\subfloat[Cost]{\includegraphics[width=0.1\linewidth]{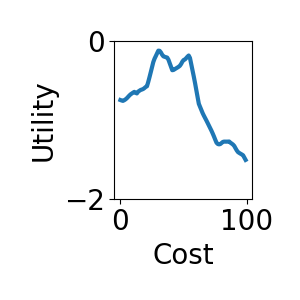}\label{sfig:cm_resnet_001_x0}}
\subfloat[IVT]{\includegraphics[width=0.1\linewidth]{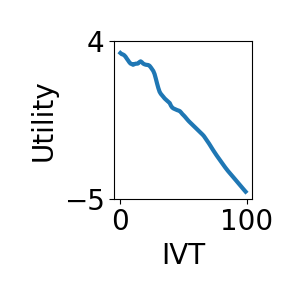}\label{sfig:cm_resnet_001_x1}}
\subfloat[Cost]{\includegraphics[width=0.1\linewidth]{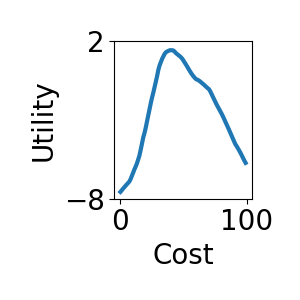}\label{sfig:cm_dnn_x0}} 
\subfloat[IVT]{\includegraphics[width=0.1\linewidth]{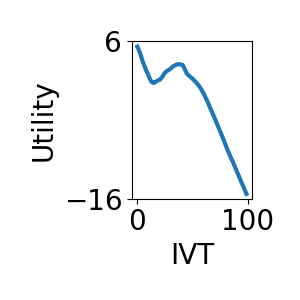}\label{sfig:cm_dnn_x1}} 
\\
\caption{Utility functions of MNL-ResNets, MNL, and DNNs. Upper row: visualization of 2D utility functions, and percentages in the parentheses represent the prediction accuracy. Lower row: visualization of 1D utility functions, and every pair of figures on the lower row correspond to the figure directly above on the upper row.}
\label{fig:interpretation_cm}
\end{figure}

\begin{figure}[H]
\captionsetup[subfigure]{justification=centering}
\centering
\subfloat[PT ($69.2\%$)]{\includegraphics[width=0.2\linewidth]{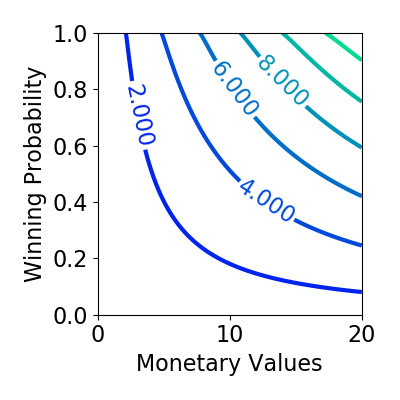}\label{sfig:pt_est_field}}
\subfloat[PT-ResNet ($\delta = 10^{-5}$; $75.6\%$)]{\includegraphics[width=0.2\linewidth]{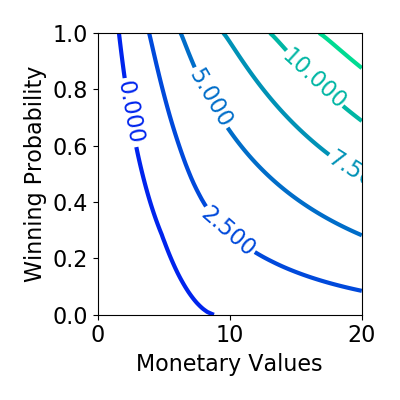}}
\subfloat[PT-ResNet ($\delta = 0.9$; $ 89.2\%$)]{\includegraphics[width=0.2\linewidth]{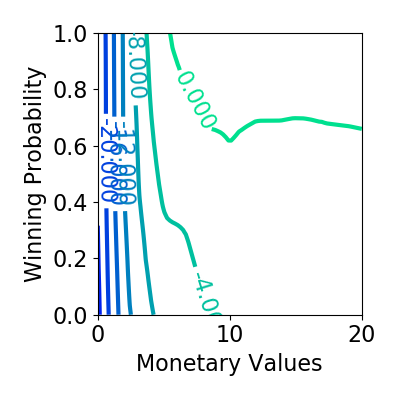}\label{sfig:pt_dnn_09_field}}
\subfloat[PT-ResNet ($\delta = 0.99$; $88.6\%$)]{\includegraphics[width=0.2\linewidth]{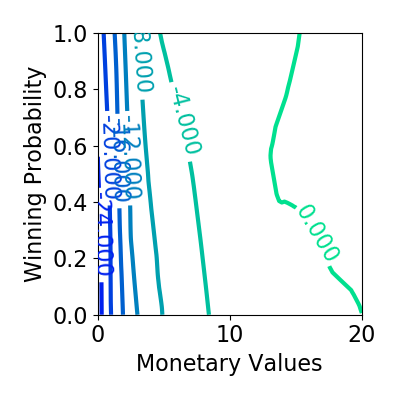}}
\subfloat[DNN ($88.3\%$)]{\includegraphics[width=0.2\linewidth]{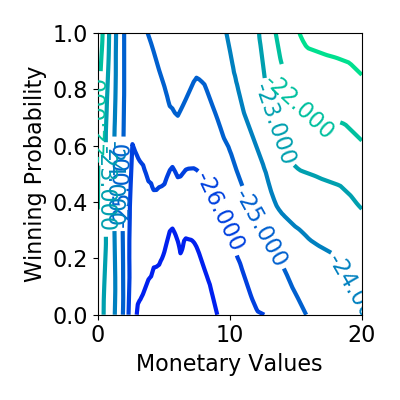}\label{sfig:pt_dnn_field}}
\\
\subfloat[Values]{\includegraphics[width=0.1\linewidth]{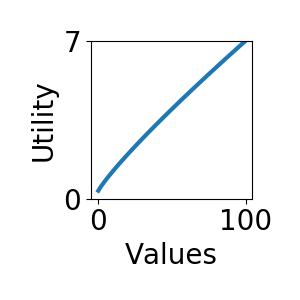}\label{sfig:pt_est_x0}}
\subfloat[Prob]{\includegraphics[width=0.1\linewidth]{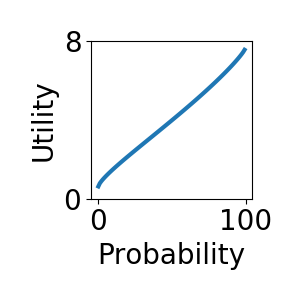}\label{sfig:pt_est_x1}}
\subfloat[Values]{\includegraphics[width=0.1\linewidth]{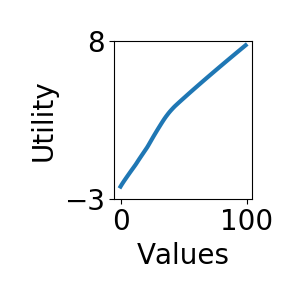}}
\subfloat[Prob]{\includegraphics[width=0.1\linewidth]{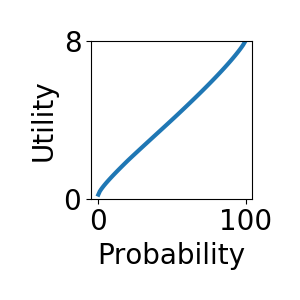}}
\subfloat[Values]{\includegraphics[width=0.1\linewidth]{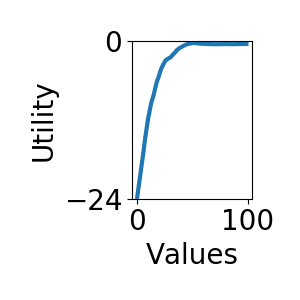}\label{sfig:pt_dnn_00001_x0}}
\subfloat[Prob]{\includegraphics[width=0.1\linewidth]{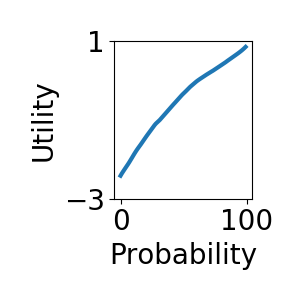}\label{sfig:pt_dnn_00001_x1}}
\subfloat[Values]{\includegraphics[width=0.1\linewidth]{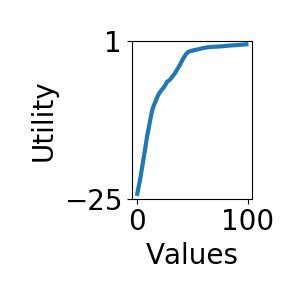}}
\subfloat[Prob]{\includegraphics[width=0.1\linewidth]{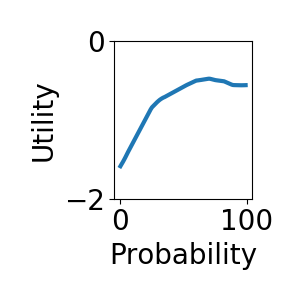}}
\subfloat[Values]{\includegraphics[width=0.1\linewidth]{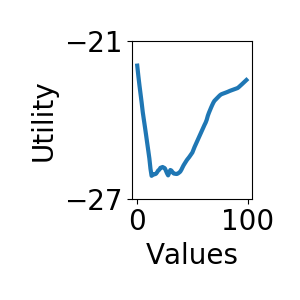}\label{sfig:pt_dnn_x0}}
\subfloat[Prob]{\includegraphics[width=0.1\linewidth]{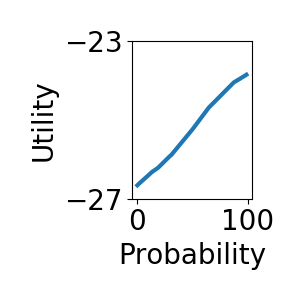}\label{sfig:pt_dnn_x1}}
\\
\caption{Utility functions of PT-ResNets, PT, and DNNs (Same format as above)}
\label{fig:interpretation_pt}
\end{figure}

\begin{figure}[H]
\captionsetup[subfigure]{justification=centering}
\centering
\subfloat[HD ($56.7\%$)]{\includegraphics[width=0.2\linewidth]{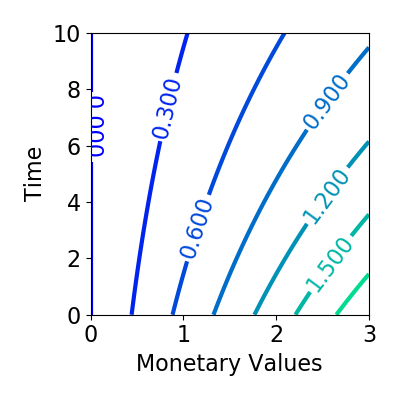}\label{sfig:hd_est_field}} 
\subfloat[HD Resnet ($\delta = 10^{-6}$; $58.0\%$)]{\includegraphics[width=0.2\linewidth]{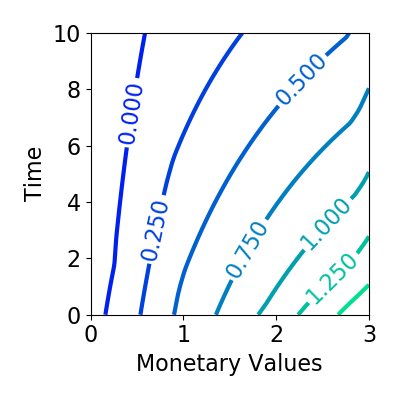}}
\subfloat[HD Resnet ($\delta = 0.05$; $77.6\%$)]{\includegraphics[width=0.2\linewidth]{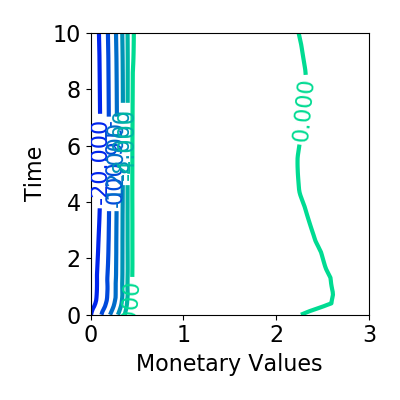}\label{sfig:hd_dnn_005_field}}
\subfloat[HD Resnet ($\delta = 0.99$; $76.0\%$)]{\includegraphics[width=0.2\linewidth]{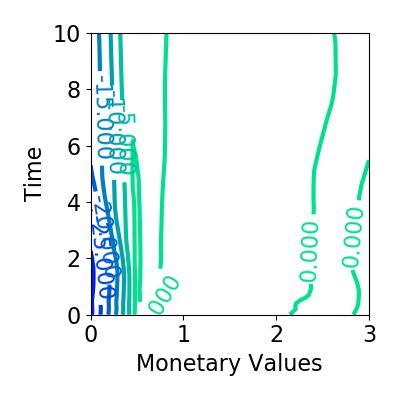}}
\subfloat[DNN ($71.2\%$)]{\includegraphics[width=0.2\linewidth]{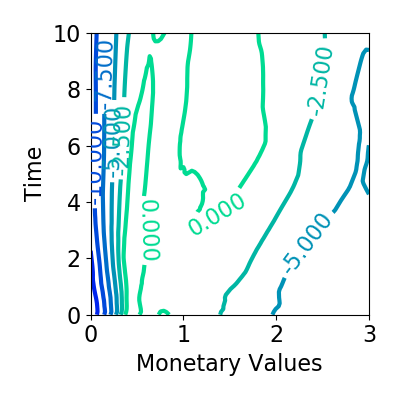}\label{sfig:hd_dnn_field}}\\
\subfloat[Values]{\includegraphics[width=0.1\linewidth]{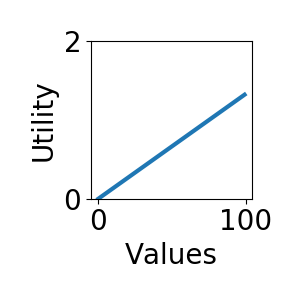}\label{sfig:hd_est_x0}}
\subfloat[Time]{\includegraphics[width=0.1\linewidth]{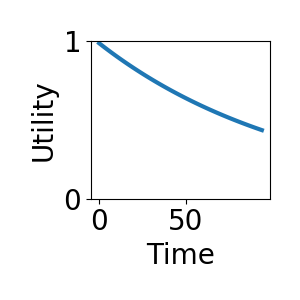}\label{sfig:hd_est_x1}}
\subfloat[Values]{\includegraphics[width=0.1\linewidth]{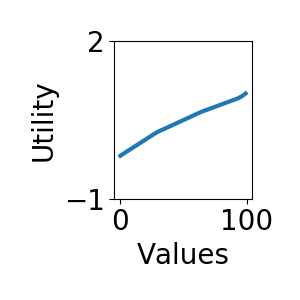}}
\subfloat[Time]{\includegraphics[width=0.1\linewidth]{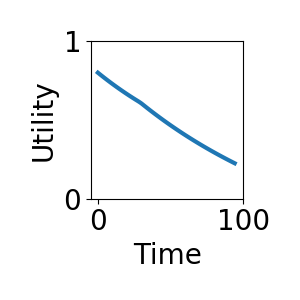}}
\subfloat[Values]{\includegraphics[width=0.1\linewidth]{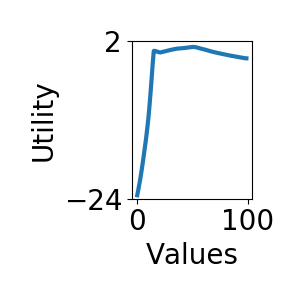}}
\subfloat[Time]{\includegraphics[width=0.1\linewidth]{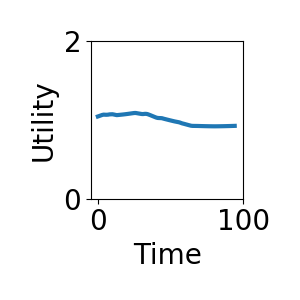}}
\subfloat[Values]{\includegraphics[width=0.1\linewidth]{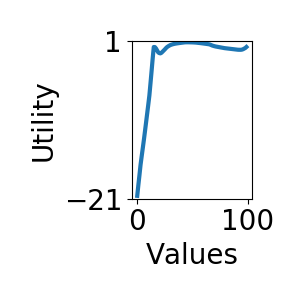}}
\subfloat[Time]{\includegraphics[width=0.1\linewidth]{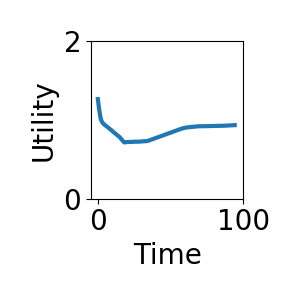}}
\subfloat[Values]{\includegraphics[width=0.1\linewidth]{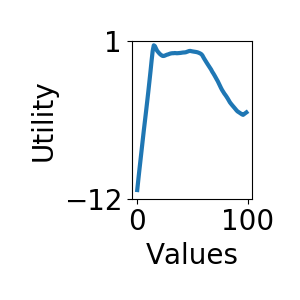}\label{sfig:hd_dnn_x0}}
\subfloat[Time]{\includegraphics[width=0.1\linewidth]{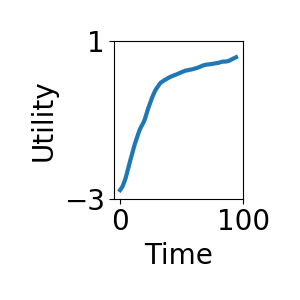}\label{sfig:hd_dnn_x1}}
\\
\caption{Utility functions of HD-ResNets, HD, and DNNs (Same format as above)}
\label{fig:interpretation_hd}
\end{figure}

\noindent
Figures \ref{fig:interpretation_cm}, \ref{fig:interpretation_pt}, and \ref{fig:interpretation_hd} visualize how utilities vary with input values in the three choice scenarios. Take Figure \ref{fig:interpretation_cm} as an example. The five graphs on the upper row represent how the utility of taking buses varies jointly with two dimensions, the monetary cost (x-axis) and the in-vehicle travel time (y-axis), and the ten graphs on the lower row visualize how utility varies with single dimensions, the monetary cost and the in-vehicle travel time respectively, holding all the other variables constant. Each pair of graphs on the lower row correspond to the one graph directly above them on the upper row. On the upper row of Figure \ref{fig:interpretation_cm}, the figure on the right end visualizes the utility function of DNN; on the left is the utility function of MNL; the three in the middle visualize the utility functions of MNL-ResNets with different $\delta$ values. The format of Figures \ref{fig:interpretation_pt} and \ref{fig:interpretation_hd} is similar to Figure \ref{fig:interpretation_cm}. In Figure \ref{fig:interpretation_pt} (PT), the monetary payoff and winning probabilities are the x- and y-axes on the upper row and the x-axes on the lower row. In Figure \ref{fig:interpretation_hd} (HD), the monetary payoff and time are the x- and y-axes on the upper row and the x-axis on the lower row. All the models in the three figures are trained by using the sequential training method and with the three-layer DNN part. The results of simultaneous training are attached in Appendix III and that of the five-layer DNN are attached in Appendix V. Both yield findings similar to the results in this subsection. 

%%%%% First, illustrate the problems of DNNs and DCMs. 
First, we can observe the complementary nature of DNNs and DCMs by comparing only the two graphs of DNNs and DCMs on the right and left ends of Figures \ref{fig:interpretation_cm}, \ref{fig:interpretation_pt}, and \ref{fig:interpretation_hd}. On one hand, the utility functions of the MNL, PT, and HD models are very regular and intuitive, as shown by subfigures \ref{sfig:cm_est_field}, \ref{sfig:pt_est_field}, and \ref{sfig:hd_est_field}. In subfigures \ref{sfig:cm_est_x0} and \ref{sfig:cm_est_x1}, the utility values of choosing the bus mode linearly decrease as bus costs and in-vehicle travel time increase. In subfigures \ref{sfig:pt_est_x0} and \ref{sfig:pt_est_x1}, the utility of taking the risky alternative increases as the monetary payoff and winning probabilities increase. These highly regular utility functions in DCMs are interpretable, although it is also likely that the true utility functions can be much more complex than the smooth and regular MNL, PT, and HD, leading to their misspecification errors and underfitting. However, on the other hand, the utility functions of DNNs for the MNL, PT, and HD scenarios are very irregular and highly counter-intuitive, although they have higher prediction accuracy, as shown by subfigures \ref{sfig:cm_dnn_field}, \ref{sfig:pt_dnn_field}, and \ref{sfig:hd_dnn_field}. For example, in Figure \ref{sfig:cm_dnn_x0}, the DNN predicts that the utility of using buses first increases as the travel cost increases, violating the basic principle of economics theory. The same type of counter-intuitive results also arise from DNNs in the PT and HD scenarios, as shown in subfigures \ref{sfig:pt_dnn_field} and \ref{sfig:hd_dnn_field}. These highly irregular utility functions in DNNs are not very interpretable, although the overly complex functions of DNNs may capture more behavioral mechanisms than DCMs, leading to higher prediction accuracy. In fact, DNNs do outperform DCMs in all three scenarios, as the prediction accuracy of DNN in the MNL setting is $55.8\%$, higher than $50.6\%$ of MNL; that of DNN in the PT setting is $88.3\%$, higher than $69.2\%$ of PT; that of DNN in the HD setting is $71.2\%$, higher than $56.7\%$ of HD. Overall, it is critical to observe the complementary nature of DNNs and DCMs: DCMs might be too simple and regular to capture reality, while DNNs might be too complex and irregular to do so.

TB-ResNets achieve a flexible compromise between DCMs and DNNs, the degree of which is controlled by $\delta$. Take the MNL setting (Figure \ref{fig:interpretation_cm}) as an example. As $\delta$ increases, the utility function of MNL-ResNets becomes more irregular and similar to the DNN model, and as $\delta$ decreases, the utility function becomes more regular and thus similar to the MNL model. This compromise also happens to the PT and HD settings: PT-ResNets and HD-ResNets resemble a continuum between the highly regular DCMs and irregular DNNs, with $\delta$ as the weighting factor. 

This perspective of TB-ResNets acting as a flexible compromise between DCMs and DNNs is tied to the six perspectives that were previously introduced. It should be obvious now that this TB-ResNet framework seeks to strike a weighted ensemble model between DCMs and DNNs through the shared utility interpretation, as shown by the continuous change in Figures \ref{fig:interpretation_cm}, \ref{fig:interpretation_pt}, and \ref{fig:interpretation_hd}. When $\delta \rightarrow 0$, the TB-ResNets increasingly resemble the theory-driven models. The DCM part becomes the skeleton utility function to stabilize the full TB-ResNet utility function, and the DNN part is strongly regularized around the DCM part. When $\delta \rightarrow 1$, the TB-ResNets can gain higher prediction accuracy, and the behavioral patterns become more similar to the irregular DNNs. Only in the middle ground, can the optimum $\delta$ values be found to construct optimum TB-ResNets.

%Different from a standard feedforward DNN, this TB-ResNet is an architecture that imposes strong regularization on the DNN part. With stronger regularizations (towards the right ends of Figures \ref{fig:interpretation_cm}, \ref{fig:interpretation_pt}, and \ref{fig:interpretation_hd}), TB-ResNets increasingly resemble DCMs. The complexity of the TB-ResNet framework is limited by two approaches: using the DCMs as the basic skeletons to stabilize the full TB-ResNet utility function, and searching around this local DCM utility specifications with the $\lambda$ regularization.

\subsection{Interpretability}
\noindent
DCMs tend to be too simple to capture reality, while DNNs tend to be too complex to do so. As a result, TB-ResNets are more interpretable than pure DCMs because TB-ResNets enrich the overly simple utility functions of DCMs with the DNN component, and also more interpretable than the pure DNNs because TB-ResNets regularize the overly complex DNNs with a small $\delta$. For example, the MNL-ResNet model ($\delta = 0.008$ in Figure \ref{sfig:cm_resnet_0008_field}) is similar to the MNL model, since it has a relatively regular utility contour and the overall utility values decrease as the cost and in-vehicle travel time increase. But unlike the MNL model, the MNL-ResNet ($\delta = 0.008$) has a richer pattern to reflect the real decision-making mechanism. On the other side, compared to pure DNNs, the MNL-ResNet ($\delta = 0.008$) retains the general decreasing trend in the utility function, capturing the pattern that the utility of taking buses should decrease with higher travel time and costs. The MNL-ResNet ($\delta = 0.008$) is more interpretable than the DNN owing to the regular local utility pattern in the base MNL component. 

PT-ResNets and HD-ResNets demonstrate the same story. An axiom underlying PT is its monotonic increasing property, since larger winning rates and monetary payoffs lead to a higher probability of choosing the alternative. However, this monotonic increasing property is violated in the DNN model, as shown in Figures \ref{sfig:pt_dnn_field}, \ref{sfig:pt_dnn_x0}, and \ref{sfig:pt_dnn_x1}, rendering the DNN model difficult to interpret. Compared to pure DNNs, the optimum PT-ResNet ($\delta = 0.9$ in Figure \ref{sfig:pt_dnn_09_field}) retains the monotonicity of the PT model and improves it by allowing a richer pattern augmented by the DNN component. In terms of the HD models, HD-ResNets reveal a similarly successful pattern, showing richer utility patterns than the pure HD and more regular patterns than the pure DNN. The utility function of the HD model is supposed to increase with higher monetary value and decrease with longer waiting time (more temporal discounting), as shown in Figures \ref{sfig:hd_est_field}, \ref{sfig:hd_est_x0}, and \ref{sfig:hd_est_x1}. The best HD-ResNet ($\delta = 0.05$ in Figure \ref{sfig:hd_dnn_005_field}) retains these overall patterns with richer details, and is more reasonable than the DNN model (Figures \ref{sfig:hd_dnn_field}, \ref{sfig:hd_dnn_x0}, and \ref{sfig:hd_dnn_x1}), which produced the counter-intuitive result that the utility of a future payoff increases with longer waiting time.

Elasticity is another important metric used to interpret choice models. Although the utility functions of DNNs and TB-ResNets are highly nonlinear, the most practical elasticity values are how the aggregate choice probabilities (market shares) respond to a 1\% change in the cost, which can simulate the policy scenarios of imposing a gasoline tax or increasing bus fares. The elasticity can be computed by the formula:
\begin{flalign}
\sum_{i = 1}^N \frac{\partial P_{i,k_1} / P_{i,k_1}}{\partial x_{i,k_2} / x_{i,k_2}} = \sum_{i} \frac{\partial P_{i,k_1}}{\partial x_{i,k_2}} \times \frac{x_{i,k_2}}{P_{i,k_1}}
\end{flalign}

\noindent in which $k_1$ and $k_2$ are the indices of two alternatives and $i$ the index of individuals. When $k_1 = k_2$, this equation computes the self-elasticity; when $k_1 \neq k_2$, it computes the cross-elasticity. With this formula, the elasticity can be computed for every output alternative regarding every input.

Table \ref{table:elasticity} summarizes the elasticity coefficients of MNL, DNNs, and MNL-ResNet with respect to the alternative-specific variables, demonstrating again that MNL-ResNets achieve a reasonable compromise between MNL and DNN. As shown in Panel 1, the elasticities of MNL follow the independence of irrelevant alternative (IIA) assumption, as the coefficients of self-elasticities are negative and that of cross-elasticities are positive. This pattern is simple and intuitive, but is often criticized as being too restrictive. On the other side, the DNN model reveals a much more irregular elasticity pattern, in which the coefficients of self-elasticities and many cross-elasticities are of a large magnitude and are negative. The elasticity coefficients in the MNL-ResNet achieve a certain compromise between MNL and DNN because the magnitude of the coefficients in MNL-ResNet is much smaller than DNNs and shrinks toward the magnitude of the MNL model, and the MNL-ResNet still retains a relatively flexible substitution pattern as in the DNN. 

\begin{table}[H]
\caption{Elasticity coefficients of MNL, DNN, and MNL-ResNet ($\delta = 0.008$)}
\centering
\resizebox{0.65\textwidth}{!}{
\begin{tabular}{l|lllll}
\hline
\textbf{Panel 1: MNL}                             & Walk   & Bus    & Ridesharing & Drive  & AV     \\ \hline
Walk: walk time              & \textbf{-1.778} & 0.127  & 0.127       & 0.127  & 0.127  \\
Bus: cost                    & 0.149  & \textbf{-0.577} & 0.149       & 0.149  & 0.149  \\
Bus: in-vehicle time         & 0.121  & \textbf{-0.431} & 0.121       & 0.121  & 0.121  \\
Ridesharing: cost            & 0.024  & 0.024  & \textbf{-0.192}      & 0.024  & 0.024  \\
Ridesharing: in-vehicle time & 0.089  & 0.089  & \textbf{-0.784}      & 0.089  & 0.089  \\
Drive: cost                  & 0.274  & 0.274  & 0.274       & \textbf{-0.760} & 0.274  \\
Drive: in-vehicle time       & 0.260  & 0.260  & 0.260       & \textbf{-0.418} & 0.260  \\
AV: cost                     & 0.044  & 0.044  & 0.044       & 0.044  & \textbf{-0.408} \\
AV: in-vehicle time          & 0.058  & 0.058  & 0.058       & 0.058  & \textbf{-0.540} \\ 
\hline
\hline
\textbf{Panel 2: DNN}                             & Walk   & Bus    & Ridesharing & Drive  & AV     \\ \hline
Walk: walk time              & \textbf{-6.298} & 1.572  & 1.022       & 0.292  & 0.597  \\
Bus: cost                    & \textbf{-0.563} & \textbf{-1.842} & \textbf{-0.193}      & 0.739  & 1.505  \\
Bus: in-vehicle time         & 0.641  & \textbf{-2.152} & \textbf{-0.494}      & 0.296  & \textbf{-0.184} \\
Ridesharing: cost            & 0.604  & 0.514  & \textbf{-2.553}      & 0.062  & 2.051  \\
Ridesharing: in-vehicle time & \textbf{-0.534} & \textbf{-0.832} & \textbf{-5.236}  & 0.553  & 0.601  \\
Drive: cost                  & 0.502  & 1.557  & 2.777       & \textbf{-2.133} & 1.984  \\
Drive: in-vehicle time       & 1.041  & 2.787  & 2.257       & \textbf{-1.643} & 3.130  \\
AV: cost                     & \textbf{-0.828} & \textbf{-0.320} & 0.841       & 0.388  & \textbf{-5.919} \\
AV: in-vehicle time          & \textbf{-0.478} & 0.752  & 0.794       & 0.041  & \textbf{-4.594} \\ 
\hline
\hline
\textbf{Panel 3: MNL-ResNet}                            & Walk   & Bus    & Ridesharing & Drive  & AV     \\ \hline
Walk: walk time              & \textbf{-2.476} & 0.478  & 0.375       & 0.189  & 0.275  \\
Bus: cost                    & \textbf{-0.262} & \textbf{-0.936} & \textbf{-0.017}      & 0.420  & 0.346  \\
Bus: in-vehicle time         & 0.363  & \textbf{-0.785} & 0.035      & 0.145  & 0.488  \\
Ridesharing: cost            & \textbf{-0.108}  & 0.099 & \textbf{-1.003}      & 0.043  & 0.627  \\
Ridesharing: in-vehicle time & \textbf{-0.117} & \textbf{-0.109} & \textbf{-1.472}     & 0.210  & 0.056  \\
Drive: cost                  & 0.283  & 0.632  & 0.708       & \textbf{-0.908} & 0.486  \\
Drive: in-vehicle time       & 0.465  & 0.802  & 0.569       & \textbf{-0.880} & 0.569  \\
AV: cost                     & \textbf{-0.010} & 0.122  & 0.481       & 0.195  & \textbf{-1.670} \\
AV: in-vehicle time          & \textbf{-0.342} & 0.232  & 0.289       & 0.190  & \textbf{-1.783} \\ \hline
\end{tabular}
}
\label{table:elasticity}
\end{table}

However, the concept of interpretability is ambiguous, since it has multiple definitions, including simulatability, decomposability, algorithmic transparency, and post-hoc interpretability \cite{Lipton2016}. Our evaluation above has defined interpretability as accurately approximating the true behavioral mechanism and being consistent with the accepted behavior knowledge \cite{WangShenhao2018_stat_learning,WangShenhao2020_econ_info}, which belongs to one type of post-hoc interpretability \cite{Ribeiro2016,Montavon2018,Hinton2015}. Mathematically, it is defined as the distance between the true and estimated choice probability functions \cite{WangShenhao2018_stat_learning}, since an accurately estimated choice probability function can provide the complete economic information in demand modeling \cite{WangShenhao2020_econ_info}. However, a different definition of interpretability can lead to a different conclusion. For example, suppose that interpretability is defined as simulatability, which means the ease with which modelers can ``simulate'' the model structure in their mind. The simplest DCMs are then fully interpretable while complex DNNs are not. The TB-ResNet, as a mixture of DCMs and DNNs, can hardly be more interpretable than simple DCMs. Although intuitive, this simulatability definition can be misleading, since it encourages an overly simplified model that fails to recognize rich behavioral reality. For example, a constant model stripped of all the contents (e.g. y = 0) can be evaluated as the most interpretable, although it is not practically useful. Nonetheless, it is important to recognize that many definitions of interpretability can co-exist, from which we adopted only a single relevant one for this work. 

\subsection{Prediction}
\begin{figure}[H]
\captionsetup[subfigure]{justification=centering}
\centering
\subfloat[MNL]{\includegraphics[width=0.33\linewidth]{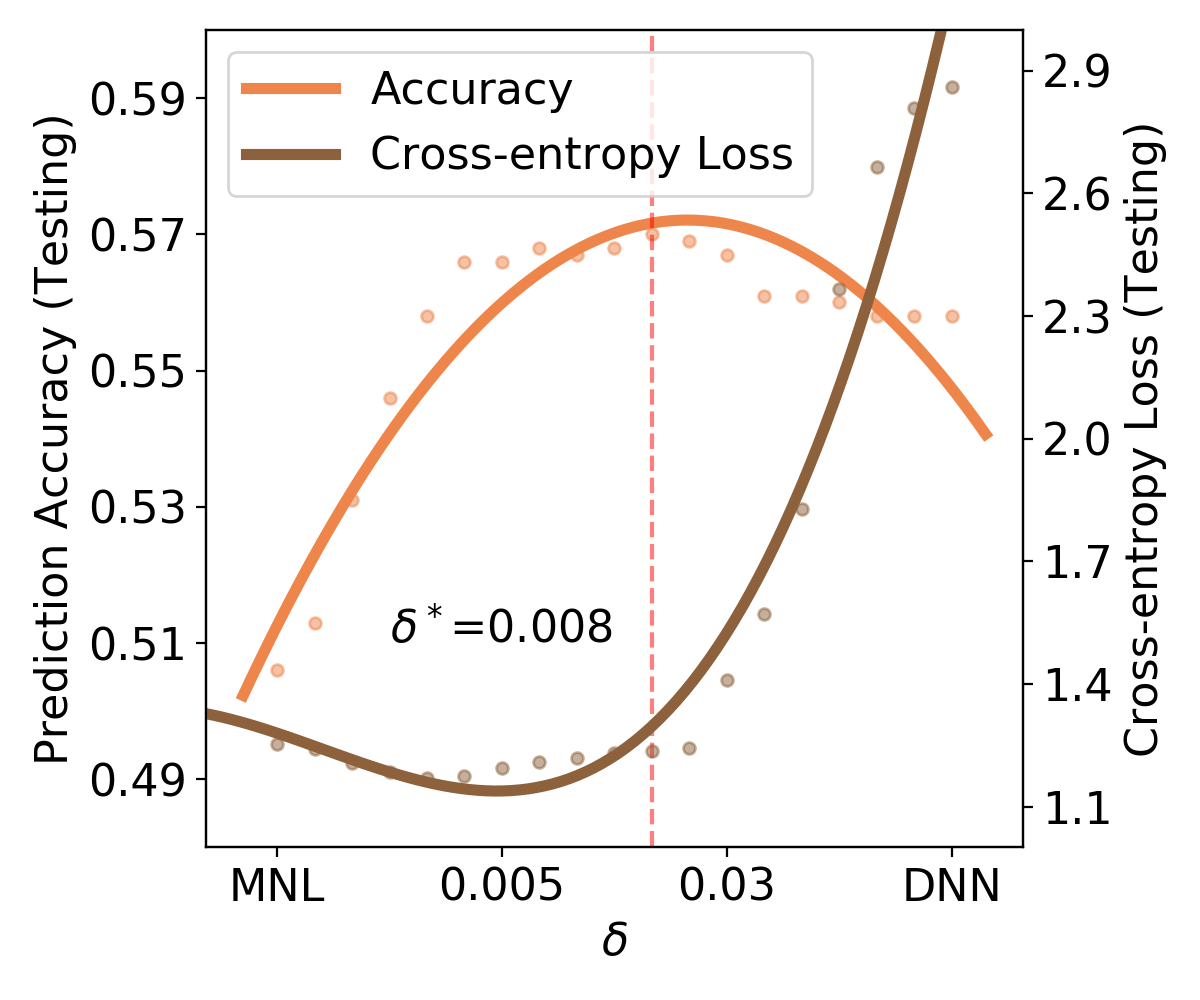}\label{sfig:cm_convergence}}
\subfloat[PT]{\includegraphics[width=0.33\linewidth]{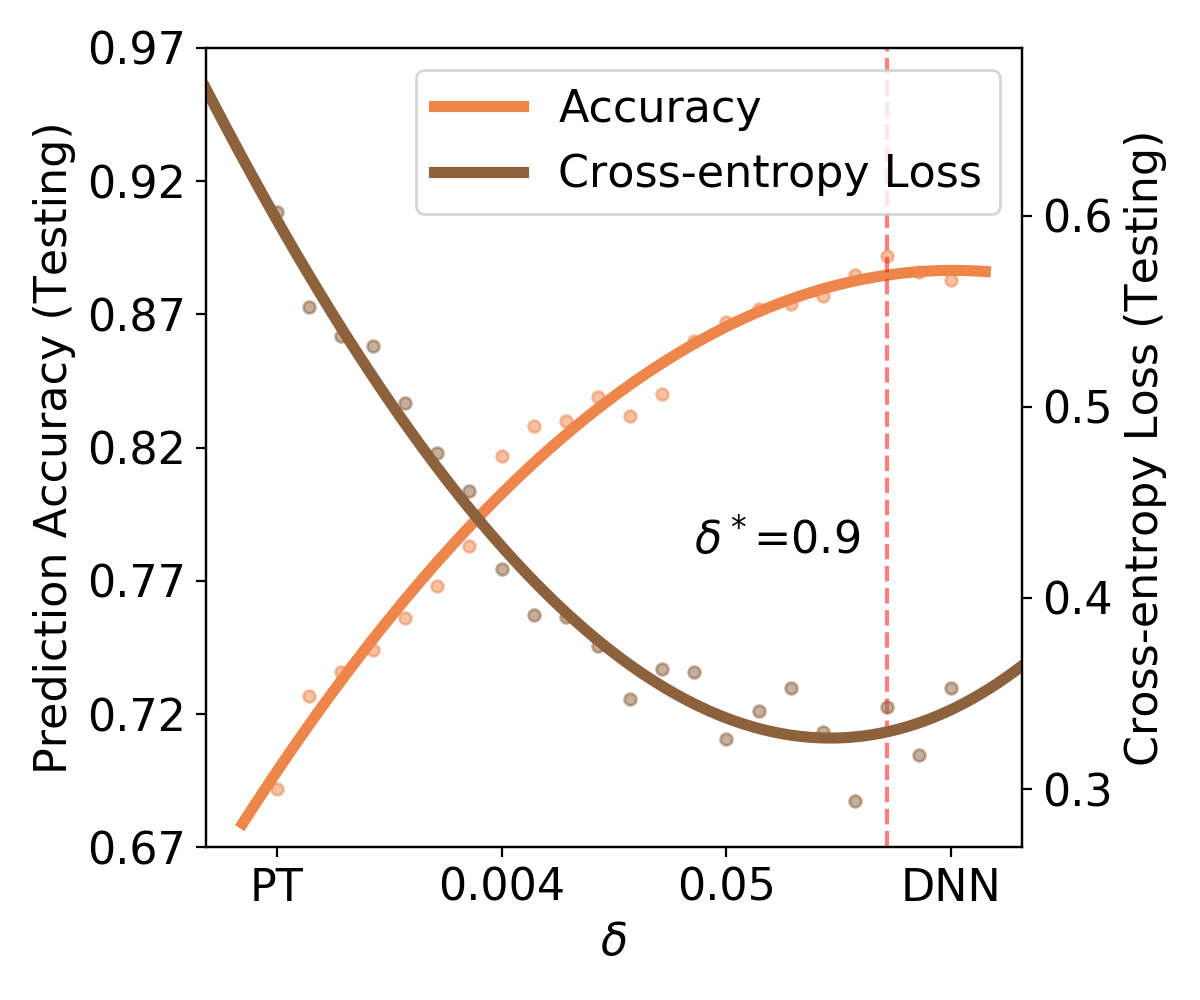}\label{sfig:pt_convergence}}
\subfloat[HD]{\includegraphics[width=0.33\linewidth]{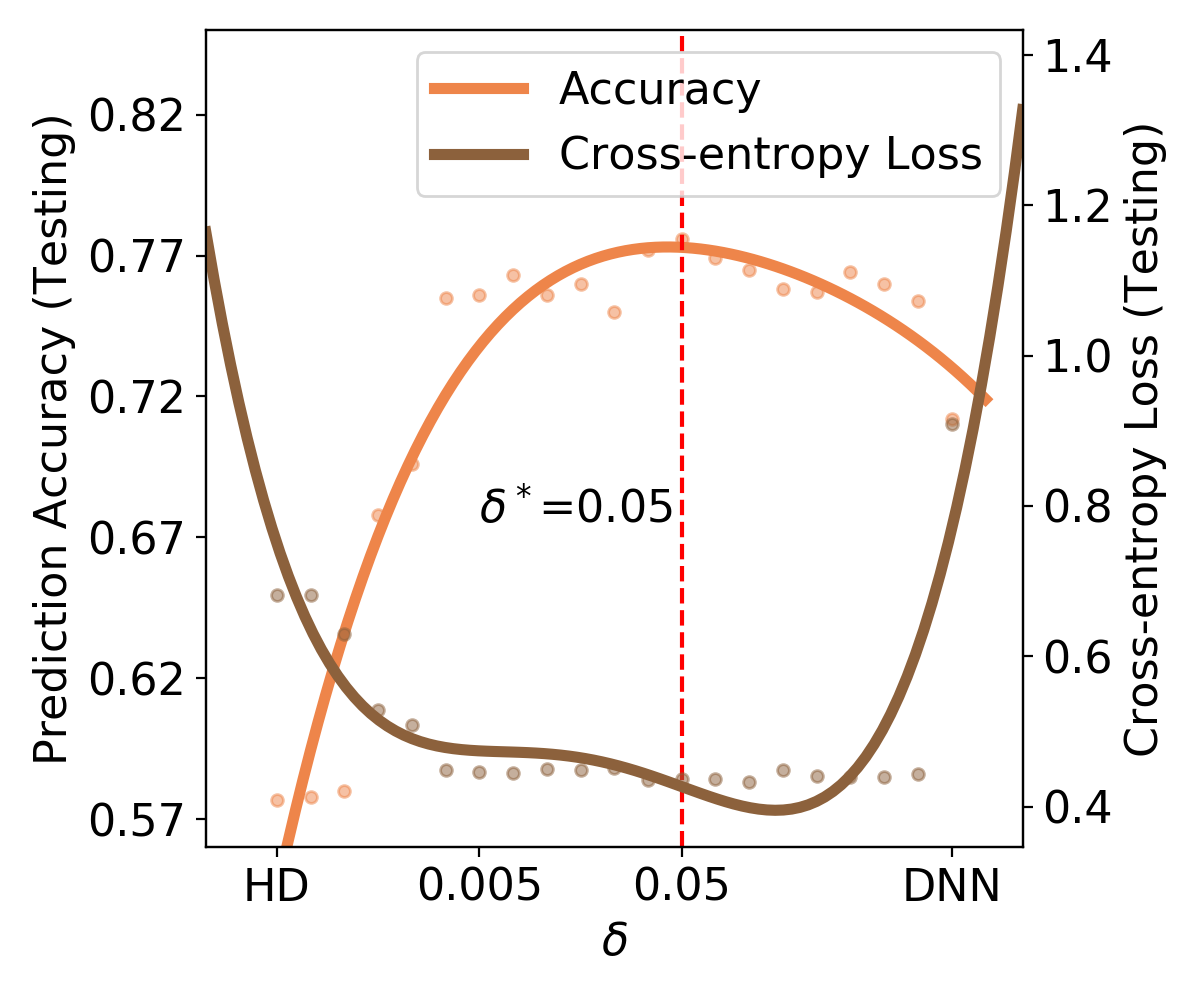}\label{sfig:hd_convergence}}
\caption{Model performance and delta (TB-ResNets with the 3-layer DNN)}
\label{fig:convergence}
\end{figure}

\noindent Table \ref{table:model_performance} summarizes the predictive performance of the three choice scenarios in three panels. It presents three metrics - prediction accuracy, cross-entropy loss, and F1 score - to measure the performance of two TB-ResNets with different DNN architectures, one with three layers and the other with five. Each panel includes the DCMs, DNNs, and TB-ResNets with varying $\delta$ values. Columns 2-6 present the prediction accuracy, cross-entropy loss, and F1 score in the testing sets of the two architectures, and the last column reports the largest share of the choice set as a baseline performance metric. To provide an intuition for the predictive performance, Figure \ref{fig:convergence} visualizes how the model performance varies with $\delta$ in the three choice scenarios, with the points representing the individual model performance, fitted by smooth curves, and the red dash lines marking the optimum $\delta$ values. 

\begin{table}[H]
\centering
\caption{Performance of DCMs, DNNs, and two TB-ResNets in testing sets (Sequential training)}
\resizebox{1.0\linewidth}{!}{
\begin{tabular}{p{0.27\linewidth}|P{0.12\linewidth}|P{0.12\linewidth}|P{0.12\linewidth}|P{0.12\linewidth} |P{0.12\linewidth}|P{0.12\linewidth}|P{0.12\linewidth}}
\toprule
& Prediction accuracy (3-layer) & Cross-entropy loss (3-layer) & F1 score (3-layer) & Prediction accuracy (5-layer) & Cross-entropy Loss (5-layer) & F1 score (5-layer) & Baseline (Largest share)  \\
\midrule
\multicolumn{8}{l}{\textbf{Panel 1. Performance of MNL models}} \\
\midrule
MNL & 50.6\% & 1.254 & 0.439 
& 50.6\% & 1.254 & 0.439& 44.8\% \\
MNL ResNet ($\delta$ = 1e-5) & 53.1\% & \textbf{1.207} & 0.485 
&52.1\% & 1.348 & 0.498 & 44.8\% \\
MNL ResNet (\textbf{$\delta$ = 0.008}) & \textbf{57.0\%} & 1.237 & \textbf{0.559}
&56.4\% & 3.118 & 0.557& 44.8\% \\
MNL ResNet ($\delta$ = 0.05) & 56.1\% & 1.572 & 0.559 
&57.8\% & 3.342 & 0.576 & 44.8\%\\
DNN & 55.8\% & 2.861 & 0.555 
& 55.3\% & 4.175 & 0.548  & 44.8\%  \\
\midrule
\multicolumn{8}{l}{\textbf{Panel 2. Performance of PT models}} \\
\midrule
PT & 69.2\% & 0.602 & 0.693 
& 69.2\% & 0.602 & 0.693 & 53.8\%\\
PT ResNet ($\delta$ = 1e-05) & 75.6\% &  0.502 & 0.762
& 81.3\% & 0.477& 0.772& 53.8\% \\
PT ResNet (\textbf{$\delta$ = 0.9}) & \textbf{89.2\%} &  0.343 & \textbf{0.887} 
&89.8\% & 0.366 & 0.900 & 53.8\%\\
PT ResNet ($\delta$ = 0.99) & 88.6\% &  \textbf{0.318} & 0.882 
&90.0\% & 0.337& 0.899& 53.8\% \\
DNN & 88.3\% & 0.353 & 0.885 
& 89.0\% & 0.368 & 0.891 & 53.8\% \\
\midrule
\multicolumn{8}{l}{\textbf{Panel 3. Performance of HD models}} \\
\midrule
HD & 56.7\% & 0.684 & 0.568 
& 56.7\% & 0.684 & 0.568& 50.0\%\\
HD ResNet ($\delta$ = 1e-05) & 68.9\% & 0.523 & 0.689 
& 73.7\% & 0.471 & 0.737& 50.0\%\\
HD ResNet (\textbf{$\delta$ = 0.05}) & \textbf{77.6\%} & \textbf{0.437} & \textbf{0.764} 
& 75.4\% & 0.445 & 0.767& 50.0\% \\
HD ResNet ($\delta$ = 0.99) & 76.0\% & 0.444 & 0.763 
& 74.9\% & 0.453 & 0.760 & 50.0\%\\
DNN & 71.2\% & 0.909 & 0.722
& 72.6\% & 0.958 & 0.726 & 50.0\%\\
\bottomrule
\end{tabular}
} %end resizing
\label{table:model_performance}
\end{table}

MNL-ResNets, PT-ResNets, and HD-ResNets outperfrom both DNNs and DCMs in prediction accuracy, cross-entropy losses, and F1 score, as shown in Table \ref{table:model_performance} and Figure \ref{fig:convergence}. In Figure \ref{fig:convergence}, the curves of the prediction accuracy are always concave and that of the cross-entropy loss are always convex. It suggests that TB-ResNets can outperform both the right and left ends in predicting individual choices, evaluated in the deterministic rule (accuracy) and the probabilistic rule (cross-entropy). The curves in Figure \ref{fig:convergence} are smooth when approaching the right and left ends, which suggest the smooth convergence of TB-ResNets towards DCMs as $\delta$ decreases and towards DNNs as $\delta$ increases. The TB-ResNet framework can outperform DCMs because the DNN part in the TB-ResNets relaxes the stringent structure constraint embedded in DCMs, which nearly always leads to misspecification errors and underfitting. The TB-ResNet framework can outperform DNNs because of the localization in the DCM part and the regularization by the small $\delta$. Overall, the results here clearly demonstrate the TB-ResNets can outperform both DCMs and DNNs, incorporate the two model families as two specific cases, and can converge to the two ends with $\delta$ approaching zero or one. 

The optimum $\delta$ values, as highlighted in bold font in Table \ref{table:model_performance}, suggest that the MNL and HD theories are more complete than the PT. As marked by the middle (3rd) model in each panel, the optimum $\delta$ values equal to 0.008 and 0.05 in the MNL and HD scenarios, while that in the PT scenario is around 0.9. As introduced in Section \ref{sec:s3_est_errors}, an optimum and small $\delta$ strongly regularizes the DNN part, suggesting relative completeness of the DCM part; while an optimum and large $\delta$ uses the DNN part to search in a larger function space, suggesting that the DCM part is less complete. Therefore, our results imply that the MNL and HD theories are relatively complete in capturing the main behavioral mechanism, while the PT has more room for improvement. This finding can also be validated by the gap of the predictive performance between DCMs and DNNs in the three scenarios. A larger prediction gap is related to a larger optimum $\delta$ value: the optimum $\delta$ values are 0.008, 0.05, and 0.9 in MNL, HD, and PT scenarios, corresponding to the prediction accuracy gaps of 5.2\% (=55.8\%-50.6\%), 14.5\% (=71.2\%-56.7\%), and 19.1\% (=88.3\%-69.2\%). Intuitively, a larger predictive gap between the DCMs and the DNNs suggests more incompleteness of the DCMs. Therefore, the empirical results highly align with our theoretical discussions. The empirical optimum $\delta$ value creates an optimum TB-ResNet model and functions as a diagnostic tool to evaluate the completeness of the theories. 

Although Table \ref{table:model_performance} presents the optimum $\delta$ as exact values, the optimum $\delta$ is more likely to take a range of values, caused by the potential inconsistency of the multiple predictive metrics \cite{WangShenhao2020_rpsp}. Model selections based on multiple predictive metrics can be inconsistent with each other. For example, in Table \ref{table:model_performance} and Figure \ref{fig:convergence}, while DNNs outperform PT in both prediction accuracy and cross-entropy loss, DNNs outperform MNL and HD models in only prediction accuracy but underperform in cross-entropy loss. In selecting the optimum TB-ResNets, the optimum $\delta$ based on the prediction accuracy and the cross-entropy loss is always slightly different. The authors would argue that this optimum range is actually more preferred to an exact optimum value, as it presents flexibility for further design of $\delta$ to enrich the TB-ResNet framework.

The TB-ResNets' predictive performance varies with the design of the DNN part. Table \ref{table:model_performance} presents the TB-ResNets constructed by two DNN parts with three and five layers, reported respectively in columns 2-4 and 5-7. The TB-ResNet with the three-layer DNN architecture slightly outperforms that with the five-layer DNN part in the MNL scenario, while the former underperforms the latter in the PT and HD scenarios. Therefore, it is possible to improve the TB-ResNets by adopting more effective DNN architecture, given a vast number of available DNN architectures. However, this work seeks to demonstrate that, regardless of the specific DNN architecture, the TB-ResNets can always improve the performance of DNNs through the combination with the DCMs. Hence, in-depth exploration into either the DNN or the DCM part is beyond the scope of our work.

\subsection{Robustness}
\noindent
Although robustness is not a traditional topic in travel demand modeling, it is increasingly important in both general ML discussions and the specific setting of demand modeling, since it can measure the local regularity of economic information. A model lacking robustness tends to predict an irregular behavioral pattern in which a slight perturbation of the input leads to a large change of the output, which is behaviorally unrealistic since a tiny price increase should not dramatically change decision-maker's choice. In practice, perturbations can arise from the measurement noises in data or the randomness in data collection process.\footnote{For more details, readers can refer to \citeauthor{WangShenhao2020_econ_info} (\citeyear{WangShenhao2020_econ_info}) to understand the relationship between robustness and local regularity of economic information.}

This type of behavioral regularity can be formally measured by robustness, or more specifically, the change of predictive performance under small perturbations of the inputs. This study uses three perturbations including two adversarial attacks (FGSM and TGSM) and one random noise (Gaussian noise). They are (1) the fast gradient sign attack method (FGSM) ($x_{adv} = x + \epsilon \times sign(\nabla_{x} L(y, \hat{y}))$), and (2) the target gradient sign method (TGSM) ($x_{adv} = x - \epsilon \times sign(\nabla_{x} L(y_{target}, \hat{y}))$) \cite{Goodfellow2015,Kurakin2016}, and (3) the Gaussian noise (GN) ($x_{adv} = x + \epsilon \times \Delta x $, where $x$ is already standardized and $ \Delta x \sim \mathcal{N}(0,1) $). The models are trained on the initial data points $x$, perturbed with the $x_{adv}$ under the three mechanisms, and evaluated by using the $x_{adv}$ to predict the $y$ corresponding to the initial $x$. A robust model should respond modestly to the perturbation, since the small input perturbations should lead to a small change of outputs. The Gaussian noise perturbation simulates random noises in data sets, while FGSM and TSGM methods simulate malicious attacks on the system. The adversarial attacks are common methods to evaluate system robustness in the ML literature, while Gaussian noise is more realistic in approximating random noise in the context of choice modeling. 

\begin{figure}[H]
\captionsetup[subfigure]{justification=centering}
\centering
\subfloat[MNL FGSM]{\includegraphics[width=0.2\linewidth]{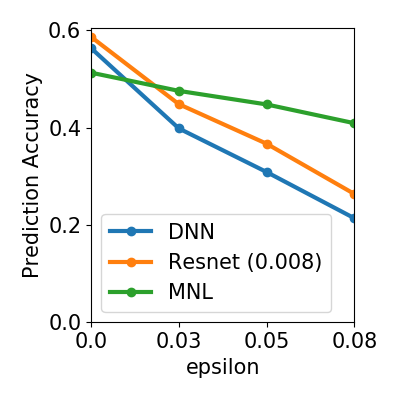}\label{sfig:cm_fgsm}}
\subfloat[PT FGSM]{\includegraphics[width=0.2\linewidth]{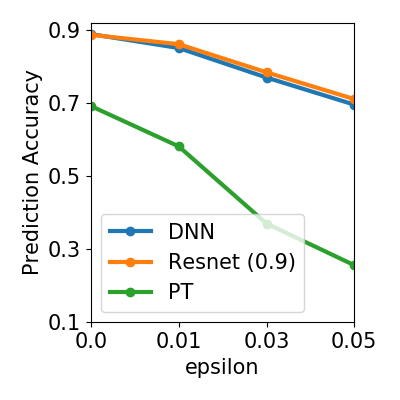}\label{sfig:pt_fgsm}}
\subfloat[HD FGSM]{\includegraphics[width=0.2\linewidth]{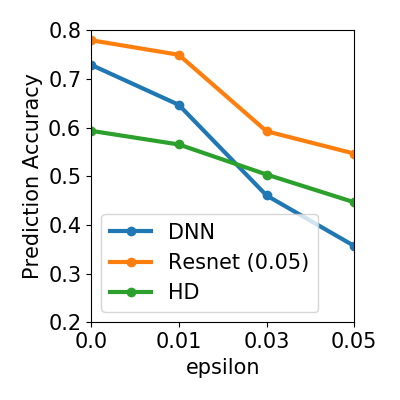}\label{sfig:hd_fgsm}}\\
\subfloat[MNL TGSM]{\includegraphics[width=0.2\linewidth]{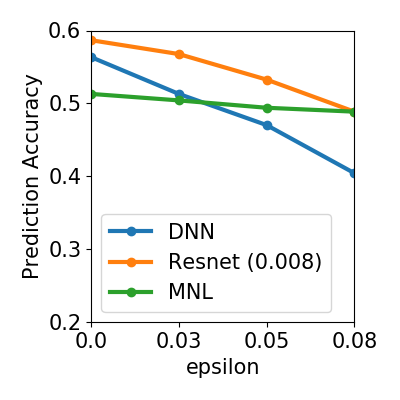}\label{sfig:cm_tgsm}}
\subfloat[PT TGSM]{\includegraphics[width=0.2\linewidth]{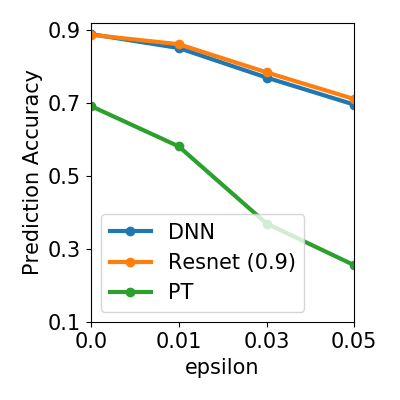}\label{sfig:pt_tgsm}}
\subfloat[HD TGSM]{\includegraphics[width=0.2\linewidth]{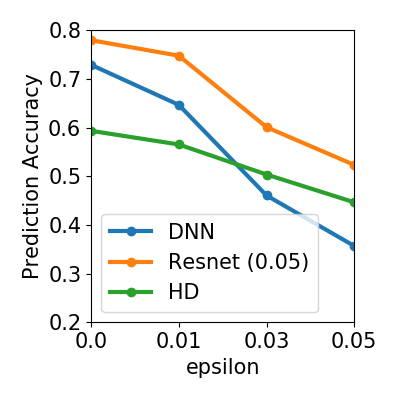}\label{sfig:hd_tgsm}}\\
\subfloat[MNL Gaussian noise]{\includegraphics[width=0.2\linewidth]{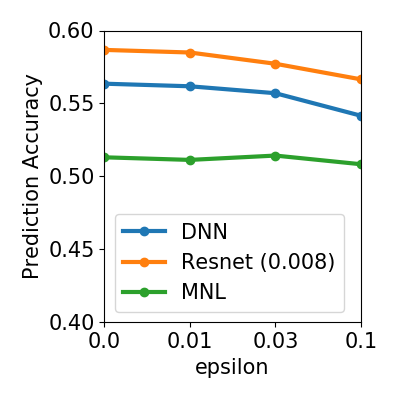}\label{sfig:cm_gn}}
\subfloat[PT Gaussian noise]{\includegraphics[width=0.2\linewidth]{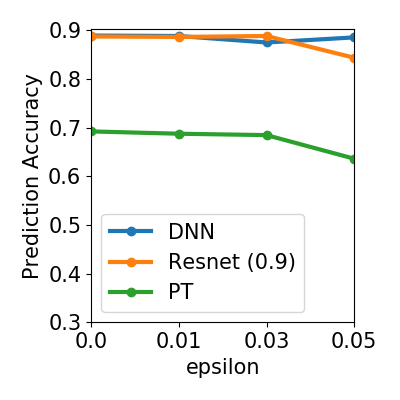}\label{sfig:pt_gn}}
\subfloat[HD Gaussian noise]{\includegraphics[width=0.2\linewidth]{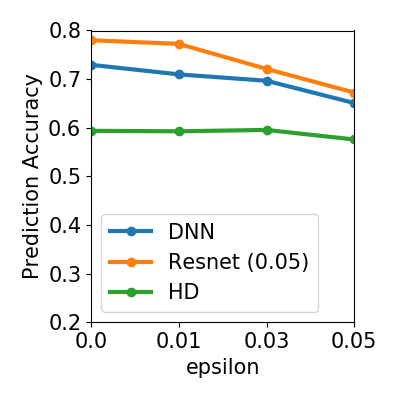}\label{sfig:hd_gn}}\\
\caption{Prediction accuracy with three perturbations (Gaussian noise, FGSM, and TGSM) using the perturbed testing sets}
\label{fig:adv_accuracy}
\end{figure}

MNL-ResNets and HD-ResNets are more robust than DNNs under the two adversarial attacks, as shown in Figures \ref{sfig:cm_fgsm}, \ref{sfig:cm_tgsm}, \ref{sfig:hd_fgsm}, and \ref{sfig:hd_tgsm}. First, DCMs are more robust than DNNs. For example, in Figure \ref{sfig:cm_fgsm}, while the prediction accuracy of the MNL model is lower than that of the DNN model when $\epsilon = 0$, the accuracy of the MNL model becomes much higher than that of the DNN when $\epsilon > 0.03$. Since the MNL-ResNet and the HD-ResNet models are the combination of DCMs and DNNs, the two TB-ResNets are more robust than DNNs, as shown by the orange curves lying above the blue curves in Figures \ref{sfig:cm_fgsm}, \ref{sfig:cm_tgsm}, \ref{sfig:hd_fgsm}, and \ref{sfig:hd_tgsm}. The reason is that the DCMs are functioning as an anchoring skeleton in the TB-ResNet system, so the local utility patterns of TB-ResNets are less irregular than DNNs. This robustness intuition is closely related to the previous discussions about utility patterns: when utility patterns are more regular, the system tends to be more robust.

However, the robustness pattern under the PT scenario is quite different from MNL and HD, because of the particularity of PT and the different optimum $\delta$ value in PT-ResNets. While MNL and HD theories are designed to have smooth and well-bounded input gradients, the probability weighting function in the PT can have an input gradient that is close to infinity when the probabilities approach zero. The underlying behavioral intuition is that people tend to significantly exaggerate very small winning chances, such as in gambling, leading to a large overestimation of the utility gain associated with the alternative. In other words, PT is designed to be sensitive to perturbations, thus lacking robustness, particularly around the region of small probabilities. This can be seen in Figures \ref{sfig:pt_fgsm} and \ref{sfig:pt_tgsm}: the PT models not only have smaller prediction accuracy than the DNNs as $\epsilon = 0$, but also decrease much more quickly than the DNNs. But on the other side, the prediction accuracy decrease of PT-ResNets and DNNs largely align with each other, because of the large optimum $\delta$ value in PT-ResNet. When $\delta$ is close to one, the PT-ResNet largely resembles the DNN part, so their robustness performance tends to be similar. Nonetheless, the pattern of an infinite gradient seems to be very specific to PT, because generally a DCM should have regular and bounded gradients, thus stabilizing the TB-ResNets rather than destabilizing them.

Under the Gaussian noise, changes in the predictive performance are much more modest, although the pattern still appears similar to that from the adversarial attacks. As shown in Figures \ref{sfig:cm_gn}, \ref{sfig:pt_gn}, and \ref{sfig:hd_gn}, all of the curves are nearly flat with slight downward sloping, suggesting that the prediction accuracy does not vary much under Gaussian noise. The difference between Gaussian noise and adversarial attacks is no surprise, since the adversarial attacks target the most vulnerable part of the input space while Gaussian noise is random. However, even with the relatively flat curves, the findings here appear similar to those from the adversarial attacks. In the MNL and HD scenarios, MNL and HD models are slightly more robust than the DNNs, while the PT model seems slightly less so than the DNNs. As a result, the MNL-ResNet and the HD-ResNet appear slightly more robust than the DNNs in the MNL and HD scenarios, while the PT-ResNet and the DNN in the PT scenario nearly overlap. 

\section{Conclusions and Discussions}
\noindent
This study introduces a TB-ResNet framework to analyze individual decision-making by synergizing the theory- and data-driven methods, based on the utility interpretation shared by DNNs and DCMs. The TB-ResNet framework can be understood from the perspectives of architecture design, model ensemble, gradient boosting, regularization, function approximation, and theory diagnosis. Three instances of TB-ResNets, including MNL-ResNets, PT-ResNets and HD-ResNets, are created and tested empirically on three data sets to evaluate model prediction, interpretability, and robustness.

As summarized in Table \ref{table:method_improve}, our empirical results demonstrate that TB-ResNets are more predictive, interpretable, and robust overall than pure DCMs and DNNs, although several exceptions exist. Compared to DNNs, TB-ResNets are more predictive, interpretable, and robust because the DCM component in TB-ResNets can stabilize the utility functions and regularize the DNN component. Compared to DCMs, TB-ResNets are more predictive and interpretable because richer utility functions are augmented to the skeleton DCM by the DNN component in TB-ResNets. The TB-ResNets are formulated with a flexible ($\delta$, $1-\delta$) weighting, thus taking advantage of both the simplicity of the DCMs and the richness of the DNNs, preventing the underfitting of the DCMs and the overfitting of the DNNs, and providing insights into the completeness of the DCM theories. Our findings are consistent across the three scenarios (MNL, PT, and HD). While exceptions exist in the PT scenario and the comparison to DCMs for robustness evaluation, our main findings hold from both theoretical and empirical perspectives.

\begin{table}[htb]
\centering
\resizebox{1.0\linewidth}{!}{
\begin{tabular}{p{0.25\linewidth}|P{0.3\linewidth}|P{0.3\linewidth}|P{0.3\linewidth}}
\toprule
\textbf{Models} & \textbf{Prediction} & \textbf{Interpretability} & \textbf{Robustness} \\
\midrule
Compared to DNNs & Marginal improvement \newline (by stabilization and regularization) & Significant improvement \newline (by stabilization and regularization) & Significant improvement \newline (by stabilization and regularization) \\
\hline
Compared to DCMs & Significant improvement \newline (by augmenting and enriching utility function) & Significant improvement \newline (by augmenting and enriching utility function) & No improvement \\
\bottomrule
\end{tabular}
} %end resizing
\caption{Comparison of TB-ResNets to DCMs and DNNs}
\label{table:method_improve}
\end{table}

The TB-ResNet is a method to reconcile the handcrafted and the automatic utility specification, which resonates with the broad discussions about how to interact the general-purpose ML and domain-specific models. DNNs generate dominant predictive performance because they can automatically learn utility specification, termed as an ``end-to-end'' system that can ``learn from scratch''. This power of automation is treated as its main strength over traditional methods that rely on domain knowledge to handcraft utility functions \cite{Mullainathan2017,LeCun2015}. However on the other side, researchers argued that automatic feature learning with zero prior knowledge does not appear to be a viable approach. Liao and Poggio \cite{Qianli2018} contended that ``being lazy (automatic learning) is good, but being too lazy is not''. Our TB-ResNet is a tangible example of integrating handcrafted and automatic learning systems in the context of individual decision-making. 

This TB-ResNet framework reveals the particularity of DNNs among all the ML classifiers, since DNNs are closely related to the DCMs through the implicit utility interpretation and the probabilistic behavioral perspective. DNNs can model the choice probabilities using the Softmax activation function, and are thus highly compatible with the classical probabilistic behavioral modeling. On the contrary, many other ML classifiers adopt a deterministic approach in modeling the choice outputs, such as K nearest neighbors and support vector machines, and are therefore somewhat more difficult to synergize with DCMs. The TB-ResNets take advantage of the utility interpretation for this synergy, so this approach resonates better with the classical utility theory than the perspectives of model ensemble and gradient boosting. From the architecture design perspective, the TB-ResNet framework can be further improved by incorporating a large number of DNN architectures from the ML community \cite{HeKaiming2016,Krizhevsky2012} or designing new DNN architectures with behavioral knowledge \cite{WangShenhao2020_asu_dnn}. The architecture design perspective will become even more important when modelers start to use high-dimensional inputs such as imagery and natural language for demand modeling.

Overall, this study introduces a simple but flexible TB-ResNet framework. It is simple because it combines DCMs and DNNs neatly into the TB-ResNet architecture, and is flexible because it can incorporate any DCM as the theory-driven part and any DNN as the data-driven part. The three instances of the TB-ResNets provide evidence that the TB-ResNet framework is malleable enough to be applied to diverse decision scenarios with many benefits. Despite its strength, many questions still remain. Although the combination of the DNNs and DCMs can generate mutual benefits, the TB-ResNet is not the only possible approach to improve prediction, interpretation, and robustness, simply because a large number of other methods have been developed to improve DCMs and DNNs in each community. DCMs can be more predictive with richer utility specifications, and DNNs can be more predictive with better designs in architecture, hyperparameters, and training algorithms. DNNs can be made more interpretable and robust by adopting activation maximization (AM) \cite{Erhan2009}, LIME \cite{Ribeiro2016}, and minimax training procedures \cite{Madry2017}. Future studies can further enrich our TB-ResNet framework by using other DCMs for the DCM component and other DNN architectures for the DNN component. This study highlights a synergetic perspective and a comprehensive model evaluation based on three criteria, so future studies could take the complementarity of the data-driven and theory-driven methods beyond simple prediction comparison. We hope that this work can pave the way for future studies to create more links between the data-driven and theory-driven methods, because their complementary nature provides immense opportunities, their underlying perspectives are interwoven, and their synergy can overcome their respective weaknesses.

%Albeit its strength, many questions remain. Although DNNs and DCMs can create mutual benefits in prediction, interpretation, and robustness, the TB-ResNet is only one approach for the benefits to happen, simply because many other methods have been developed respectively to improve DCMs and DNNs in each community. DCMs can become more predictive with richer utility specifications, and DNNs can be more predictive with better designs in architectures, hyperparameters, and training algorithms. DNNs can be more interpretable and robust by adopting activation maximization (AM) \cite{Erhan2009}, LIME \cite{Ribeiro2016}, and minimax training procedures \cite{Madry2017}. Among the three aspects, interpretability is the most ambiguous concept. This study adopts a rather narrow perspective by focusing on only the economic interpretation through the utility function; otherwise it would be difficult to provide a high-level comparison. But readers should note that other meanings of interpretability also exist, which can lead to other modeling and practical innovations.

%To make the DNNs more robust, researchers also designed defense algorithms, including defensive distillation \cite{Papernot2016}, adversarial training by using both clean and adversarial examples \cite{Kurakin2016}, minimax formulation of robust optimization \cite{Madry2017}, and the input gradient regularizations in training \cite{Ross2018}. 

\section*{Acknowledgement}
\noindent
The research is supported by the National Research Foundation (NRF), Prime Minister’s Office, Singapore, under CREATE programme, Singapore-MIT Alliance for Research and Technology (SMART) Centre, Future Urban Mobility (FM) IRG. We thank Nick Caros for his careful proofreading. 

\section*{Author Contributions - CRediT}
\noindent
\textbf{Shenhao Wang}: Conceptualization, Methodology, Software, Formal analysis, Investigation, Writing-Original Draft, Writing-Review \& Editing, Project administration. \textbf{Baichuan Mo}: Software, Data Curation, Visualization. \textbf{Jinhua Zhao}: Supervision, Funding acquisition, Resources. All authors discussed the results and contributed to the final manuscript. 

\printbibliography

\newpage
\section*{Appendix I: Proof of Propositions 1 and 2}
\noindent
\textbf{Proposition 1 Proof.} This proof can be found in many textbooks \cite{Train2009,Ben_Akiva1985}. With Gumbel distributional assumption, Equation \ref{eq:prob_1} could be solved in an analytical way:

\begin{equation}
\setlength{\jot}{2pt} \label{eq:choice_prob_deriv}
  \begin{aligned}
  P_{ik} &= \int_{- \infty}^{+\infty} \underset{j \neq k}{\prod} e^{-e^{-(V_{ik} - V_{ij} + \epsilon_{ik})}} f(\epsilon_{ik})d\epsilon_{ik} \\
         &= \int \underset{j}{\prod} e^{-e^{-(V_{ik} - V_{ij} + \epsilon_{ik})}} e^{- \epsilon_{ik}} d\epsilon_{ik} \\
         &= \int exp(e^{- \epsilon_{ik}} \underset{j}{\sum} {-e^{-(V_{ik} - V_{ij})}}) e^{- \epsilon_{ik}} d\epsilon_{ik} \\
         &= \int_{\infty}^{0} exp(-t \underset{j}{\sum} e^{-(V_{ik} - V_{ij})} ) dt \\
         &= \frac{e^{V_{ik}}}{\underset{j}{\sum} e^{V_{ij}}}
  \end{aligned}
\end{equation}

\noindent 
in which the fourth equation uses $t = e^{- \epsilon_{ik}}$. Note this formula in Equation \ref{eq:choice_prob_deriv} is the Softmax function in DNN. $V_{ik}$ is both the deterministic utility in RUM and the inputs into the Softmax function in DNN. \\ \par

\noindent \textbf{Proposition 2 Proof.} The detailed proof of Proposition 2 could be found in Lemma 2 of McFadden's seminal paper \cite{McFadden1974}. Here is a brief summary. Suppose that one individual $i$ firstly chooses between alternative $k$ and $T$ alternatives $j$. Then according to Equations \ref{eq:prob_1} and \ref{eq:choice_prob_deriv}, 

\begin{equation}
\setlength{\jot}{2pt} \label{eq:choice_ik_1}
  \begin{aligned}
  P_{ik} &= \frac{e^{V_{ik}}}{e^{V_{ik}} + T e^{V_{ij}}} \\
         &= \int F(\epsilon_{ik} + V_{ik} - V_{ij})^T d F(\epsilon_{ik})
  \end{aligned}
\end{equation}

\noindent Suppose that the individual $i$ chooses between alternatives $k$ and alternative $l$ in another choice scenario, and alternative $l$ is constructed such that $T e^{V_{ij}} = e^{V_{il}}$. Then

\begin{equation}
\setlength{\jot}{2pt} \label{eq:choice_ik_2}
  \begin{aligned}
  P_{ik} &= \frac{e^{V_{ik}}}{e^{V_{ik}} + e^{V_{il}}} \\
         &= \int F(\epsilon_{ik} + V_{ik} - V_{il}) d F(\epsilon_{ik}) \\
         &= \int F(\epsilon_{ik} + V_{ik} - V_{ij} - log T) d F(\epsilon_{ik}) 
  \end{aligned}
\end{equation}

\noindent By construction, Equations \ref{eq:choice_ik_1} and \ref{eq:choice_ik_2} are equivalent
$$ \int F(\epsilon_{ik} + V_{ik} - V_{ij} - log T) - F(\epsilon_{ik} + V_{ik} - V_{ij})^T d F(\epsilon_{ik})  = 0 $$

\noindent Since $F(\epsilon)$ is transition complete, meaning that $\forall a$, $E h(\epsilon + a) = 0$ implies $h(\epsilon) = 0, \forall \epsilon$, it implies 

$$F(V_{ik} - log \ T) = F(V_{ik})^T, \forall V_{ik}, T$$

\noindent Taking $V_{ik} = 0$ implies $F(-log \ T) = e^{-\alpha T}$. Taking $V_{ik} = log \ T - log \ L$ implies $F(-log \ L) = F(log \ T/L)^T$. Hence $F(log \ T/L) = F(-log \ L)^{1/T} = e^{- \alpha L/T}$. Therefore, $F(\epsilon) = e^{-\alpha e^{-\epsilon}}$. This is the function of Gumbel distribution when $\alpha = 1$.

\section*{Appendix II: Summary Statistics of Three Data Sets}
\label{appendix:summary_stat}
\begin{table}[H]
\caption{Summary Statistics of SG data set}
\centering
\resizebox{\textwidth}{!}{
\begin{tabular}{llllll}
\hline
\multicolumn{3}{l|}{\textbf{Variables}}                                                                & \multicolumn{3}{l}{\textbf{Variables}}                                                               \\ \hline
Name                            & Mean                   & \multicolumn{1}{l|}{Std.}                   & Name                                                & Mean                   & Std.                  \\\hline
Male (Yes = 1)                  & 0.383                  & \multicolumn{1}{l|}{0.486}                  & Age \textless 35 (Yes = 1)                          & 0.329                  & 0.470                 \\
Age\textgreater{}60 (Yes = 1)   & 0.075                  & \multicolumn{1}{l|}{0.263}                  & Low education (Yes = 1)                             & 0.331                  & 0.471                 \\
High education (Yes = 1)        & 0.480                  & \multicolumn{1}{l|}{0.500}                  & Low income (Yes = 1)                                & 0.035                  & 0.184                 \\
High income (Yes = 1)           & 0.606                  & \multicolumn{1}{l|}{0.489}                  & Full job (Yes = 1)                                  & 0.602                  & 0.490                 \\
Walk: walk time (min)           & 60.50                  & \multicolumn{1}{l|}{54.88}                  & Bus: cost (\$SG)                                    & 2.070                  & 1.266                 \\
Bus: walk time (min)            & 11.96                  & \multicolumn{1}{l|}{10.78}                  & Bus: waiting time (min)                             & 7.732                  & 5.033                 \\
Bus: in-vehilce time (min)      & 25.06                  & \multicolumn{1}{l|}{18.91}                  & RideSharing: cost (\$SG)                            & 14.48                  & 11.64                 \\
RideSharing: waiting time (min) & 7.108                  & \multicolumn{1}{l|}{4.803}                  & RideSharing: in-vehilce time (min)                  & 18.28                  & 13.39                 \\
AV: cost (\$SG)                 & 16.08                  & \multicolumn{1}{l|}{14.60}                  & AV: waiting time (min)                              & 7.249                  & 5.674                 \\
AV: in-vehilce time (min)       & 20.11                  & \multicolumn{1}{l|}{16.99}                  & Drive: cost (\$SG)                                  & 10.49                  & 10.57                 \\
Drive: walk time (min)          & 3.968                  & \multicolumn{1}{l|}{4.176}                  & Drive: in-vehilce time (min)                        & 17.43                  & 14.10                 \\ \hline
\multicolumn{6}{l}{\textbf{Statitics}}                                                                                                                                                                        \\ \hline
Number of samples               & \multicolumn{5}{l}{8418}                                                                                                                                                    \\
Number of choices               & \multicolumn{5}{l}{\begin{tabular}[c]{@{}l@{}}Walk: 874 (10.38\%); Bus: 1951 (23.18\%); RideSharing: 904 (10.74\%);\\ Drive 3774 (44.83\%); AV: 915 (10.87\%)\end{tabular}} \\ \hline
\end{tabular}}
\end{table}

\begin{table}[H]
\caption{Summary Statistics of PT data set}
\centering
\resizebox{\textwidth}{!}{
\begin{tabular}{llllll}
\hline
\multicolumn{3}{l|}{\textbf{Variables}}                                   & \multicolumn{3}{l}{\textbf{Variables}}                     \\ \hline
Name                                 & Mean  & \multicolumn{1}{l|}{Std.}  & Name                                      & Mean   & Std.  \\ \hline
Male (Yes = 1)                       & 0.619 & \multicolumn{1}{l|}{0.485} & Age                                       & 47.46  & 12.89 \\
Num of years in school               & 6.746 & \multicolumn{1}{l|}{3.821} & Household annual income (1 million dong)  & 20.27  & 21.15 \\
Chinese (Yes = 1)                    & 0.055 & \multicolumn{1}{l|}{0.228} & Distance to the nearest local market (km) & 1.482  & 1.840 \\
Living in Southern Vietnam (Yes = 1) & 0.541 & \multicolumn{1}{l|}{0.498} & Reward 1 in option A (1 million dong)         & 0.032  & 0.016 \\
Prob of reward 1 in option A         & 0.638 & \multicolumn{1}{l|}{0.263} & Reward 2 in option A (1 million dong)         & 0.016  & 0.015 \\
Prob of reward 2 in option A         & 0.362 & \multicolumn{1}{l|}{0.263} & Reward 1 in option B (1 million dong)         & 0.076  & 0.038 \\
Prob of reward 1 in option B         & 0.486 & \multicolumn{1}{l|}{0.252} & Reward 2 in option B (1,000 dong)         & -0.340 & 9.640 \\
Prob of reward 2 in option B         & 0.514 & \multicolumn{1}{l|}{0.252} &                                           &        &       \\ \hline
\multicolumn{6}{l}{\textbf{Statistics}}                                                                                                 \\ \hline
Number of samples                    & \multicolumn{5}{l}{5249}                                                                        \\
Number of choices                    & \multicolumn{5}{l}{Option A: 2823 (53.78\%); Option B: 2426 (46.22\%) }             \\ \hline
\end{tabular}}
\end{table}

\begin{table}[H]
\caption{Summary Statistics of HD data set}
\centering
\resizebox{\textwidth}{!}{
\begin{tabular}{llllll}
\hline
\multicolumn{3}{l|}{\textbf{Variables}}                                                     & \multicolumn{3}{l}{\textbf{Variables}}                      \\ \hline
Name                                                   & Mean  & \multicolumn{1}{l|}{Std.}  & Name                                        & Mean  & Std.  \\ \hline
Male (Yes = 1)                                         & 0.618 & \multicolumn{1}{l|}{0.486} & Age                                         & 47.51 & 12.94 \\
Num of years in school                                 & 6.764 & \multicolumn{1}{l|}{3.843} & Household income (1 million dong)           & 20.71 & 21.23 \\
Chinese (Yes = 1)                                      & 0.055 & \multicolumn{1}{l|}{0.228} & Distance to the nearest local market (km)   & 1.506 & 1.846 \\
Living in Southern Vietnam (Yes = 1)                   & 0.534 & \multicolumn{1}{l|}{0.499} & \footnotemark Trusted agent  (Yes = 1)                 & 0.028 & 0.165 \\
Payment received in the risk experiment (1 million dong) & 20.97 & \multicolumn{1}{l|}{21.17} & Amount of immediate reward (1 million dong) & 0.075 & 0.078 \\
Amount of delayed reward (1 million dong)              & 0.150 & \multicolumn{1}{l|}{0.104} & Days of delay                               & 35.67 & 32.33 \\ \hline
\multicolumn{6}{l}{\textbf{Statistics}}                                                                                                                    \\ \hline
Number of samples                                      & \multicolumn{5}{l}{5340}                                                                         \\
Number of choices                                      & \multicolumn{5}{l}{Immediate reward: 2670 (50.0\%); Future reward: 2670 (50.0\%)}                \\ \hline
\end{tabular}
}
\end{table}
\footnotetext{Trusted agents are people who would keep the money until delayed delivery date to ensure subjects believed the money would be delivered. The selected trusted persons were usually village heads or presidents of women’s associations}

\section*{Appendix III: Results of Simultaneous Training}
\begin{figure}[H]
\captionsetup[subfigure]{justification=centering}
\centering
\subfloat[MNL FGSM]{\includegraphics[width=0.2\linewidth]{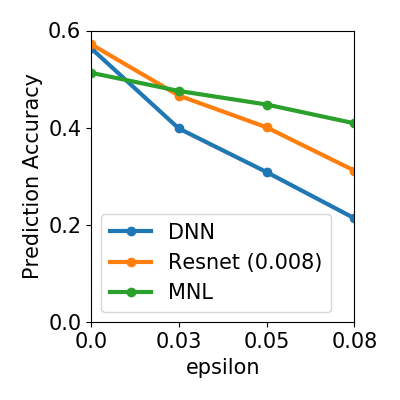}\label{sfig:cm_fgsm_append}}
\subfloat[PT FGSM]{\includegraphics[width=0.2\linewidth]{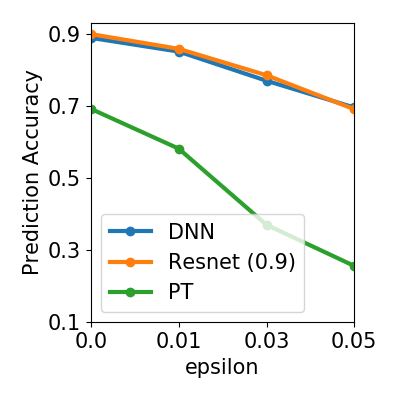}\label{sfig:pt_fgsm_append}}
\subfloat[HD FGSM]{\includegraphics[width=0.2\linewidth]{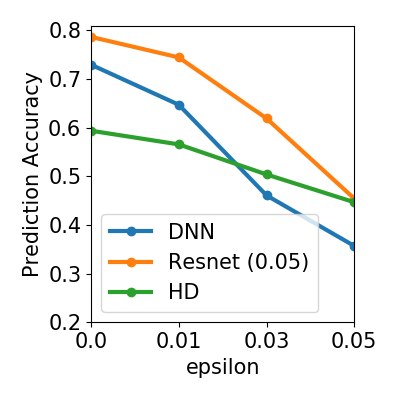}\label{sfig:hd_fgsm_append}}\\
\subfloat[MNL TGSM]{\includegraphics[width=0.2\linewidth]{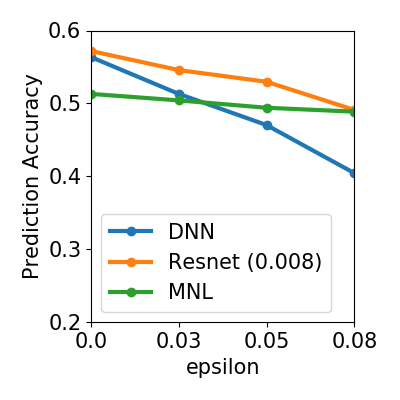}\label{sfig:cm_tgsm_append}}
\subfloat[PT TGSM]{\includegraphics[width=0.2\linewidth]{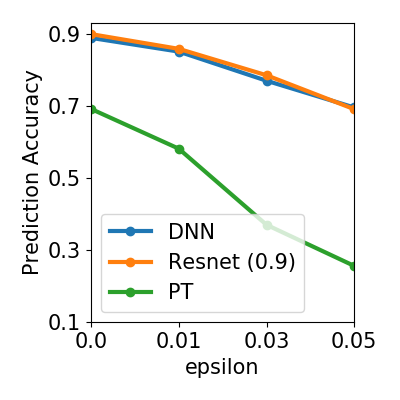}\label{sfig:pt_tgsm_append}}
\subfloat[HD TGSM]{\includegraphics[width=0.2\linewidth]{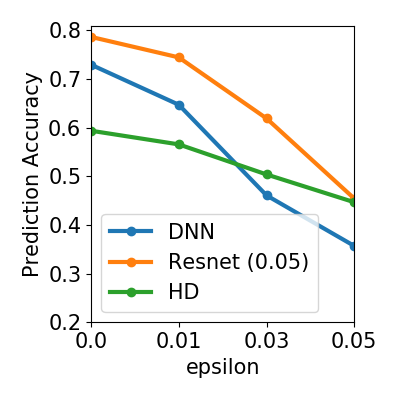}\label{sfig:hd_tgsm_append}}\\
\subfloat[MNL Gaussian noise]{\includegraphics[width=0.2\linewidth]{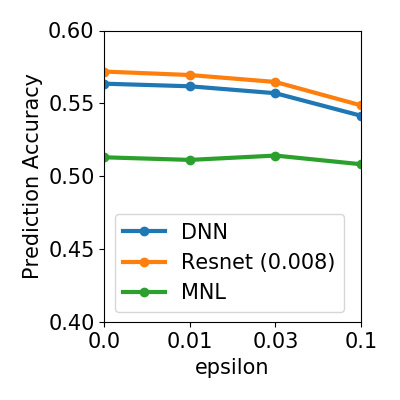}\label{sfig:cm_fgsm_append}}
\subfloat[PT Gaussian noise]{\includegraphics[width=0.2\linewidth]{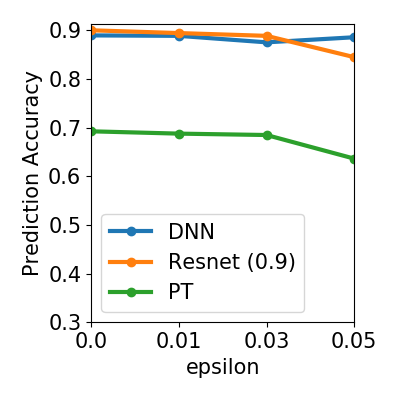}\label{sfig:pt_fgsm_append}}
\subfloat[HD Gaussian noise]{\includegraphics[width=0.2\linewidth]{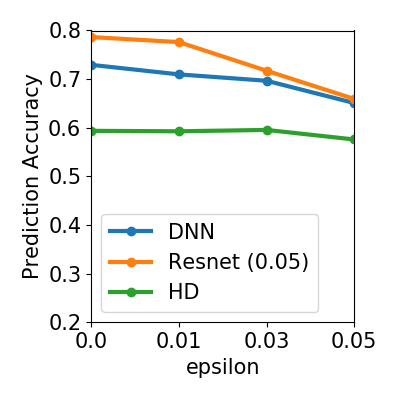}\label{sfig:hd_fgsm_append}}\\
\caption{Prediction accuracy with perturbations (Gaussian noise, FGSM, and TGSM attacks, simultaneous training)}
\label{fig:adv_accuracy_append}
\end{figure}

\begin{figure}[H]
\captionsetup[subfigure]{justification=centering}
\centering
\subfloat[MNL $(50.6\%)$]{\includegraphics[width=0.2\linewidth]{exp/interpretation/cm_est_field_use_delta.png}\label{sfig:cm_est_field_append}} 
\subfloat[MNL-ResNet ($\delta = 10^{-5}$; $52.1\%$)]{\includegraphics[width=0.2\linewidth]{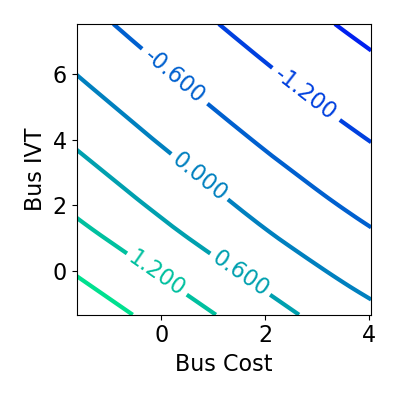}\label{sfig:cm_resnet_1e-5_field_append}}
\subfloat[MNL-ResNet ($\delta = 0.008$; $56.6\%$)]{\includegraphics[width=0.2\linewidth]{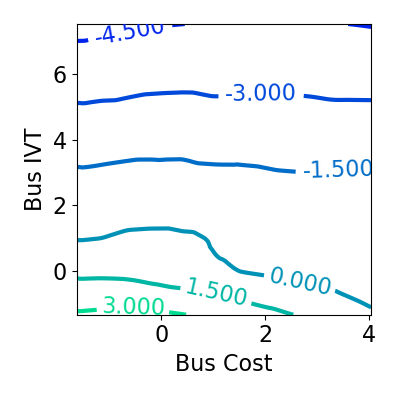}\label{sfig:cm_resnet_0005_field_append}}
\subfloat[MNL-ResNet ($\delta = 0.05$; $56.1\%$)]{\includegraphics[width=0.2\linewidth]{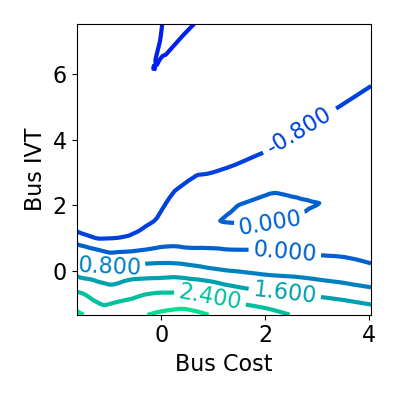}\label{sfig:cm_resnet_001_field_append}}
\subfloat[DNN ($55.8\%$)]{\includegraphics[width=0.2\linewidth]{exp/interpretation/cm_dnn_field_use_delta.png}\label{sfig:cm_dnn_field_append}} \\
\subfloat[Cost]{\includegraphics[width=0.1\linewidth]{exp/interpretation/cm_est_x0_use_delta.png}\label{sfig:cm_est_x0_append}} 
\subfloat[IVT]{\includegraphics[width=0.1\linewidth]{exp/interpretation/cm_est_x1_use_delta.png}\label{sfig:cm_est_x1_append}}
\subfloat[Cost]{\includegraphics[width=0.1\linewidth]{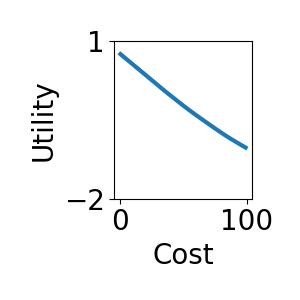}\label{sfig:cm_resnet_1e-5_x0_append}}
\subfloat[IVT]{\includegraphics[width=0.1\linewidth]{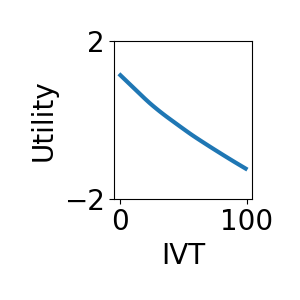}\label{sfig:cm_resnet_1e-5_x1_append}} 
\subfloat[Cost]{\includegraphics[width=0.1\linewidth]{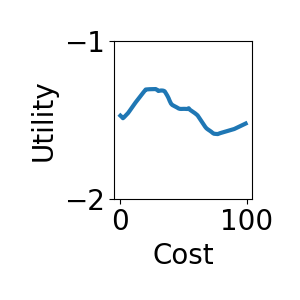}\label{sfig:cm_resnet_0005_x0_append}}
\subfloat[IVT]{\includegraphics[width=0.1\linewidth]{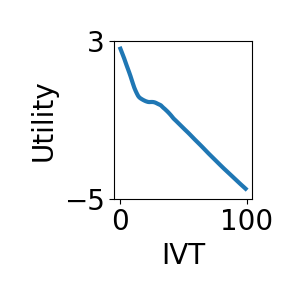}\label{sfig:cm_resnet_0005_x1_append}}
\subfloat[Cost]{\includegraphics[width=0.1\linewidth]{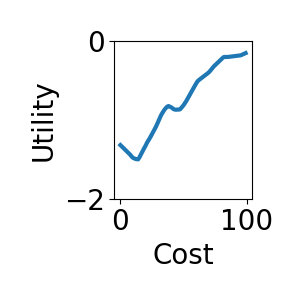}\label{sfig:cm_resnet_001_x0_append}}
\subfloat[IVT]{\includegraphics[width=0.1\linewidth]{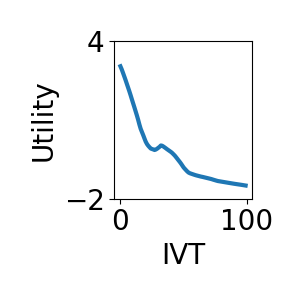}\label{sfig:cm_resnet_001_x1_append}}
\subfloat[Cost]{\includegraphics[width=0.1\linewidth]{exp/interpretation/cm_dnn_x0_use_delta.png}\label{sfig:cm_dnn_x0_append}} 
\subfloat[IVT]{\includegraphics[width=0.1\linewidth]{exp/interpretation/cm_dnn_x1_use_delta.png}\label{sfig:cm_dnn_x1_append}} 
\\
\caption{Utility functions of MNL-ResNets, MNL, and DNNs (Simultaneous training).}
\label{fig:interpretation_cm_append}
\end{figure}

\begin{figure}[H]
\captionsetup[subfigure]{justification=centering}
\centering
\subfloat[PT ($69.2\%$)]{\includegraphics[width=0.2\linewidth]{exp/interpretation/pt_est_field_use_delta.png}\label{sfig:pt_est_field_append}}
\subfloat[PT-ResNet ($\delta = 10^{-5}$; $76.7\%$)]{\includegraphics[width=0.2\linewidth]{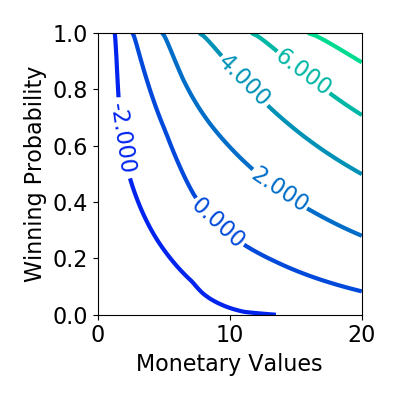}}
\subfloat[PT-ResNet ($\delta = 0.9$; $ 88.7\%$)]{\includegraphics[width=0.2\linewidth]{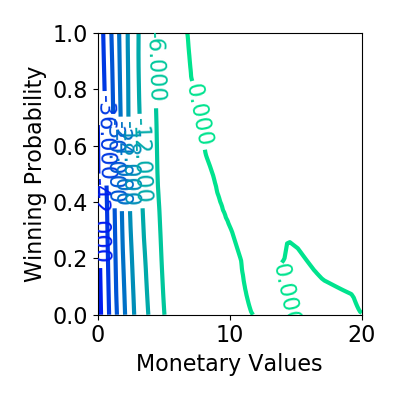}\label{sfig:pt_dnn_00001_field_append}}
\subfloat[PT-ResNet ($\delta = 0.99$; $88.6\%$)]{\includegraphics[width=0.2\linewidth]{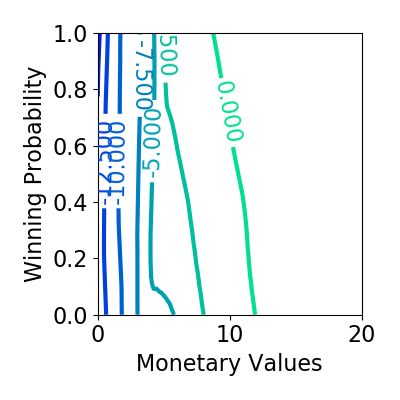}}
\subfloat[DNN ($88.3\%$)]{\includegraphics[width=0.2\linewidth]{exp/interpretation/pt_dnn_field_use_delta.png}\label{sfig:pt_dnn_field_append}}
\\
\subfloat[Values]{\includegraphics[width=0.1\linewidth]{exp/interpretation/pt_est_x0_use_delta.png}\label{sfig:pt_est_x0_append}}
\subfloat[Prob]{\includegraphics[width=0.1\linewidth]{exp/interpretation/pt_est_x1_use_delta.png}\label{sfig:pt_est_x1_append}}
\subfloat[Values]{\includegraphics[width=0.1\linewidth]{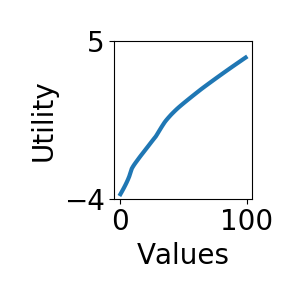}}
\subfloat[Prob]{\includegraphics[width=0.1\linewidth]{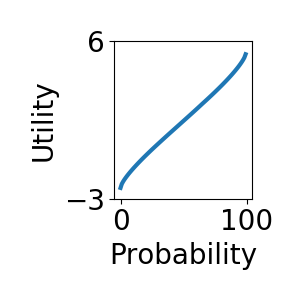}}
\subfloat[Values]{\includegraphics[width=0.1\linewidth]{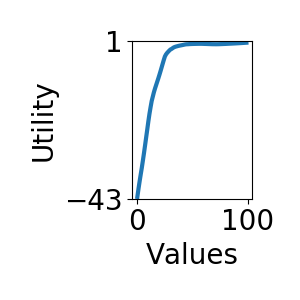}\label{sfig:pt_dnn_00001_x0_append}}
\subfloat[Prob]{\includegraphics[width=0.1\linewidth]{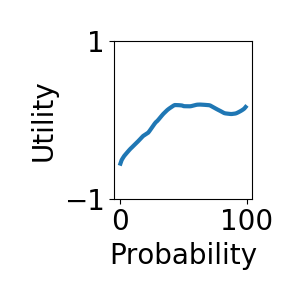}\label{sfig:pt_dnn_00001_x1_append}}
\subfloat[Values]{\includegraphics[width=0.1\linewidth]{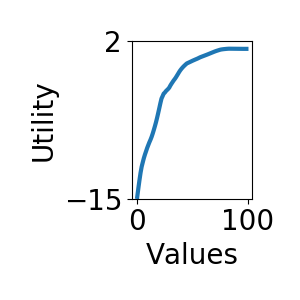}}
\subfloat[Prob]{\includegraphics[width=0.1\linewidth]{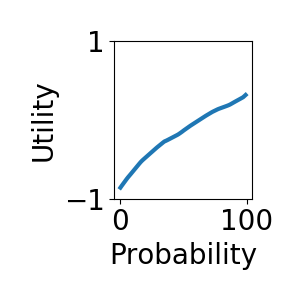}}
\subfloat[Values]{\includegraphics[width=0.1\linewidth]{exp/interpretation/pt_dnn_x0_use_delta.png}\label{sfig:pt_dnn_x0_append}}
\subfloat[Prob]{\includegraphics[width=0.1\linewidth]{exp/interpretation/pt_dnn_x1_use_delta.png}\label{sfig:pt_dnn_x1_append}}
\\
\caption{Utility functions of PT-ResNets, PT, and DNNs (Simultaneous training).}
\label{fig:interpretation_pt_append}
\end{figure}

\begin{table}[H]
\centering
\caption{Performance of DCMs, DNNs, and TB-ResNets in testing sets (Sequential and simultaneous training)}
\resizebox{1.0\linewidth}{!}{
\begin{tabular}{p{0.25\linewidth}|P{0.15\linewidth}|P{0.15\linewidth}|P{0.15\linewidth}|P{0.15\linewidth} |P{0.15\linewidth}|P{0.15\linewidth}|P{0.15\linewidth}}
\toprule
& Prediction accuracy (Sequential) & Cross-entropy loss (Sequential) & F1 score (Sequential) & Prediction accuracy (Simultaneous) & Cross-entropy Loss (Simultaneous) & F1 score (Simultaneous) & Baseline (Largest share) \\
\midrule
\multicolumn{8}{l}{\textbf{Panel 1. Performance of MNL models}} \\
\midrule
MNL & 50.6\% & 1.254 & 0.439 
& 50.6\% & 1.254 & 0.439& 44.8\% \\
MNL ResNet ($\delta$ = 1e-5) & 53.1\% & 1.207 & 0.485 
&52.1\% & 1.224 & 0.468& 44.8\% \\
MNL ResNet ($\delta$ = 0.008) & 57.0\% & 1.237 & 0.559
&56.6\% & 1.150 & 0.542& 44.8\% \\
MNL ResNet ($\delta$ = 0.05) & 56.1\% & 1.572 & 0.559 
&56.1\% & 1.213 & 0.550 & 44.8\%\\
DNN & 55.8\% & 2.861 & 0.555 
& 55.8\% & 2.861 & 0.555  & 44.8\%  \\
\midrule
\multicolumn{8}{l}{\textbf{Panel 2. Performance of PT models}} \\
\midrule
PT & 69.2\% & 0.602 & 0.693 
& 69.2\% & 0.602 & 0.693 & 53.8\%\\
PT ResNet ($\delta$ = 1e-05) & 75.6\% &  0.502 & 0.762
& 76.7\% & 0.477& 0.772& 53.8\% \\
PT ResNet ($\delta$ = 0.9) & 89.2\% &  0.343 & 0.887 
&88.7\% & 0.347 & 0.882& 53.8\%\\
PT ResNet ($\delta$ = 0.99) & 88.6\% &  0.318 & 0.882 
&88.6\% & 0.335& 0.884& 53.8\% \\
DNN & 88.3\% & 0.353 & 0.885 
& 88.3\% & 0.353 & 0.885 & 53.8\% \\
\midrule
\multicolumn{8}{l}{\textbf{Panel 3. Performance of HD models}} \\
\midrule
HD & 56.7\% & 0.684 & 0.568 
& 56.7\% & 0.684 & 0.568& 50.0\%\\
HD ResNet ($\delta$ = 1e-05) & 68.9\% & 0.523 & 0.689 
& 68.7\% & 0.517 & 0.686& 50.0\%\\
HD ResNet ($\delta$ = 0.05) & 77.6\% & 0.437 & 0.764 
& 77.3\% & 0.439 & 0.774& 50.0\% \\
HD ResNet ($\delta$ = 0.99) & 76.0\% & 0.444 & 0.763 
& 76.3\% & 0.440 & 0.774& 50.0\%\\
DNN & 71.2\% & 0.909 & 0.722 
& 71.2\% & 0.909 & 0.722 & 50.0\%\\
\bottomrule
\end{tabular}
} %end resizing
\label{table:model_performance_append}
\end{table}

\begin{figure}[!t]
\captionsetup[subfigure]{justification=centering}
\centering
\subfloat[HD ($56.7\%$)]{\includegraphics[width=0.2\linewidth]{exp/interpretation/hd_est_field_use_delta.png}\label{sfig:hd_est_field_append}} 
\subfloat[HD Resnet ($\delta = 10^{-6}$; $57.7\%$)]{\includegraphics[width=0.2\linewidth]{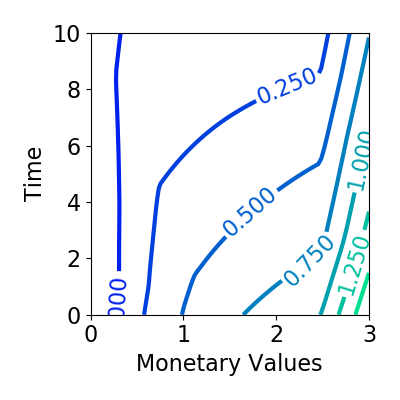}}
\subfloat[HD Resnet ($\delta = 0.05$; $77.3\%$)]{\includegraphics[width=0.2\linewidth]{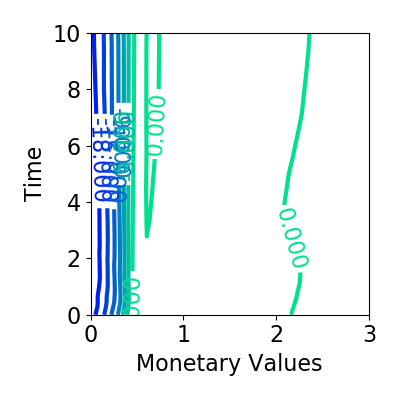}}
\subfloat[HD Resnet ($\delta = 0.99$; $76.3\%$)]{\includegraphics[width=0.2\linewidth]{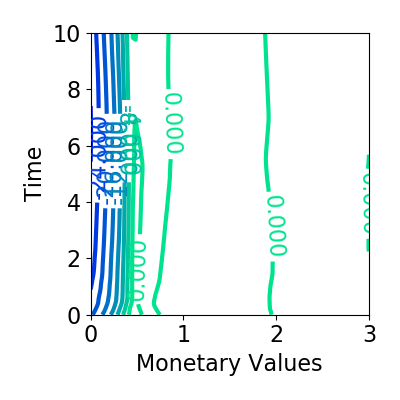}}
\subfloat[DNN ($71.2\%$)]{\includegraphics[width=0.2\linewidth]{exp/interpretation/hd_dnn_field_use_delta.png}\label{sfig:hd_dnn_field_append}}\\
\subfloat[Values]{\includegraphics[width=0.1\linewidth]{exp/interpretation/hd_est_x0_use_delta.png}\label{sfig:hd_est_x0_append}}
\subfloat[Time]{\includegraphics[width=0.1\linewidth]{exp/interpretation/hd_est_x1_use_delta.png}\label{sfig:hd_est_x1_append}}
\subfloat[Values]{\includegraphics[width=0.1\linewidth]{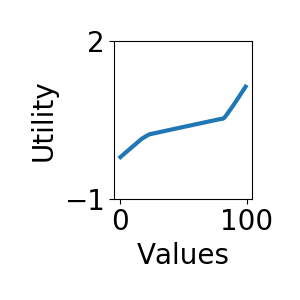}}
\subfloat[Time]{\includegraphics[width=0.1\linewidth]{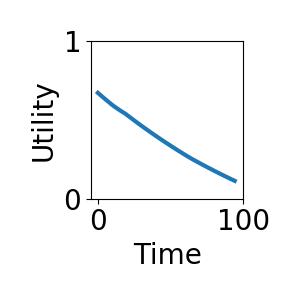}}
\subfloat[Values]{\includegraphics[width=0.1\linewidth]{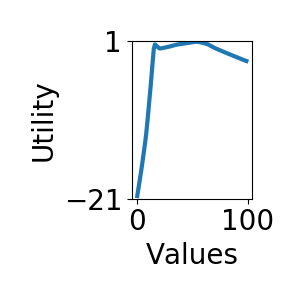}}
\subfloat[Time]{\includegraphics[width=0.1\linewidth]{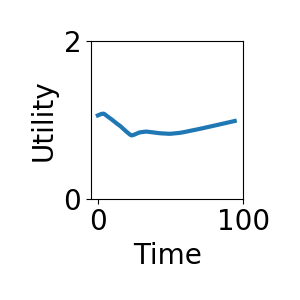}}
\subfloat[Values]{\includegraphics[width=0.1\linewidth]{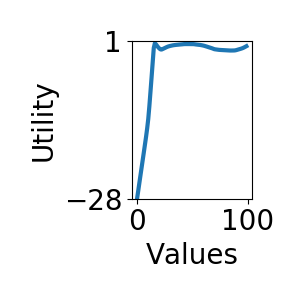}}
\subfloat[Time]{\includegraphics[width=0.1\linewidth]{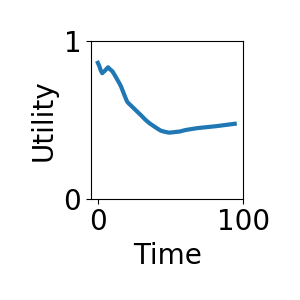}}
\subfloat[Values]{\includegraphics[width=0.1\linewidth]{exp/interpretation/hd_dnn_x0_use_delta.png}\label{sfig:hd_dnn_x0_append}}
\subfloat[Time]{\includegraphics[width=0.1\linewidth]{exp/interpretation/hd_dnn_x1_use_delta.png}\label{sfig:hd_dnn_x1_append}}
\\
\caption{Utility functions of HD-ResNets, HD, and DNNs (Simultaneous training)}
\label{fig:interpretation_hd_append}
\end{figure}

\section*{Appendix IV: Proof of Proposition 4}
\noindent \textbf{Proof.} By using the definition of Rademacher complexity and Proposition \ref{prop:3}, the left hand side can be rewritten as:
\begin{flalign}
{\E}_S[L(\hat f) - L(f^*_{\F})] & \leq2{\E}_S \hat{\R}_n({\F}|_S) \\
    & = 2{\E}_S \hat{\R}_n((1-\delta){\F}_1 + \delta {\F}_2 |_S) \\
    & = 2{\E}_S {\E}_{\epsilon} \sup_{f_1 \in {\F}_1; f_2 \in {\F}_2} \frac{1}{N} \langle \epsilon, (1-\delta)f_1 + \delta f_2 \rangle \\ 
    & \leq  2{\E}_S \big[ {\E}_{\epsilon} \sup_{f_1 \in {\F}_1} \frac{1}{N} \langle \epsilon, (1-\delta)f_1 \rangle + {\E}_{\epsilon} \sup_{f_2 \in {\F}_2} \frac{1}{N} \langle \epsilon, \delta f_2 \rangle \big] \\ 
    & = 2{\E}_S [(1-\delta)\hat{\R}_n({\F}_1|_S) + \delta \hat{\R}_n({\F}_2|_S)]
\end{flalign}

\noindent in which the first line uses Proposition \ref{prop:3}; the second line uses the definition of $\F$; the third line uses a definition $f := (1-\delta)f_1 + \delta f_2$; the fourth line uses the convexity of the $\sup$ operator; the last one uses the definition of Rademacher complexity again.

\section*{Appendix V: Utility Intuition for Five-Layer DNN Architecture}
\begin{figure}[H]
\captionsetup[subfigure]{justification=centering}
\centering
\subfloat[MNL $(50.6\%)$]{\includegraphics[width=0.2\linewidth]{exp/interpretation/cm_est_field_use_delta.png}\label{sfig:cm_est_field_append}} 
\subfloat[MNL-ResNet ($\delta = 10^{-5}$; $53.1\%$)]{\includegraphics[width=0.2\linewidth]{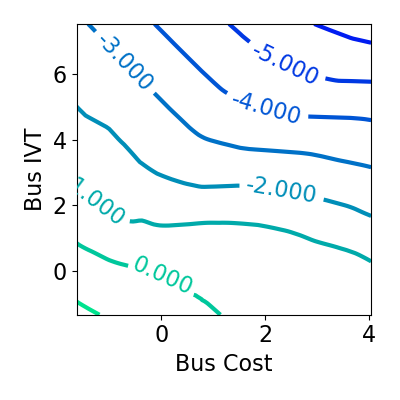}\label{sfig:cm_resnet_1e-5_field_append}}
\subfloat[MNL-ResNet ($\delta = 0.008$; $57.0\%$)]{\includegraphics[width=0.2\linewidth]{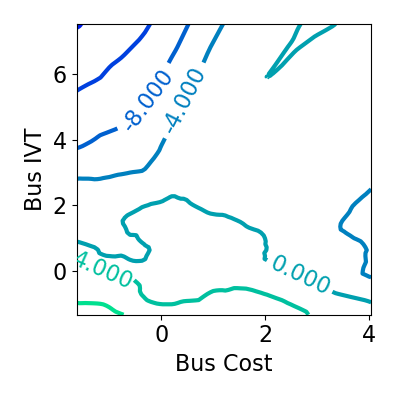}\label{sfig:cm_resnet_0005_field_append}}
\subfloat[MNL-ResNet ($\delta = 0.05$; $56.1\%$)]{\includegraphics[width=0.2\linewidth]{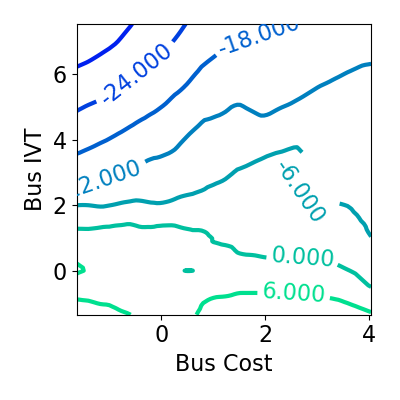}\label{sfig:cm_resnet_001_field_append}}
\subfloat[DNN ($55.8\%$)]{\includegraphics[width=0.2\linewidth]{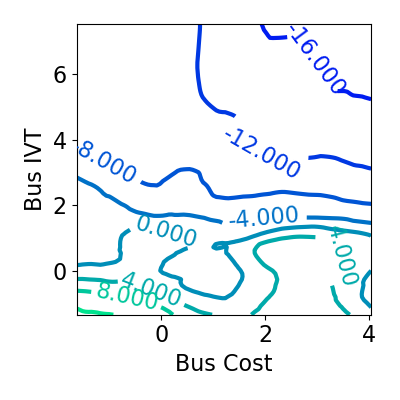}\label{sfig:cm_dnn_field_append}} \\
\subfloat[Cost]{\includegraphics[width=0.1\linewidth]{exp/interpretation/cm_est_x0_use_delta.png}\label{sfig:cm_est_x0_append}} 
\subfloat[IVT]{\includegraphics[width=0.1\linewidth]{exp/interpretation/cm_est_x1_use_delta.png}\label{sfig:cm_est_x1_append}}
\subfloat[Cost]{\includegraphics[width=0.1\linewidth]{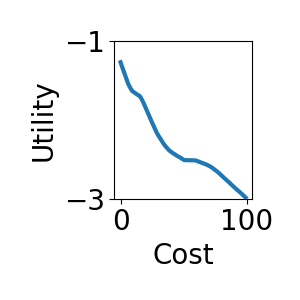}\label{sfig:cm_resnet_1e-5_x0_append}}
\subfloat[IVT]{\includegraphics[width=0.1\linewidth]{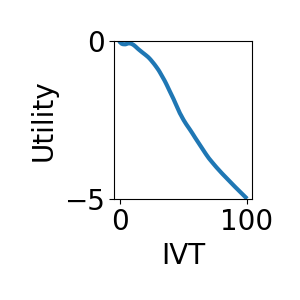}\label{sfig:cm_resnet_1e-5_x1_append}} 
\subfloat[Cost]{\includegraphics[width=0.1\linewidth]{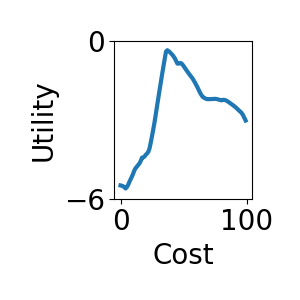}\label{sfig:cm_resnet_0005_x0_append}}
\subfloat[IVT]{\includegraphics[width=0.1\linewidth]{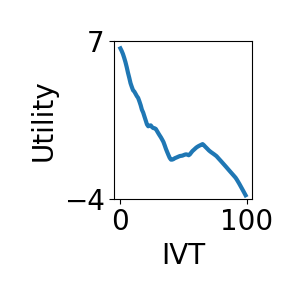}\label{sfig:cm_resnet_0005_x1_append}}
\subfloat[Cost]{\includegraphics[width=0.1\linewidth]{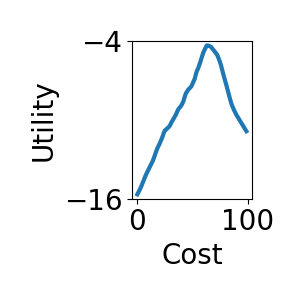}\label{sfig:cm_resnet_001_x0_append}}
\subfloat[IVT]{\includegraphics[width=0.1\linewidth]{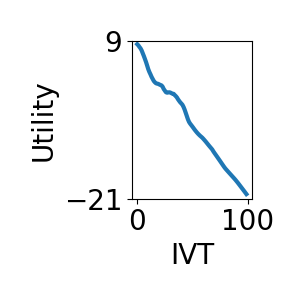}\label{sfig:cm_resnet_001_x1_append}}
\subfloat[Cost]{\includegraphics[width=0.1\linewidth]{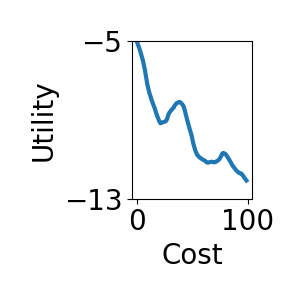}\label{sfig:cm_dnn_x0_append}} 
\subfloat[IVT]{\includegraphics[width=0.1\linewidth]{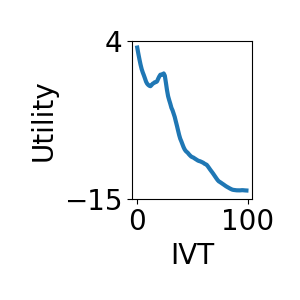}\label{sfig:cm_dnn_x1_append}} 
\\
\caption{Utility functions of MNL-ResNets, MNL, and DNNs; upper row: visualization of 2D utility functions; lower row: visualization of 1D utility functions; percentages on the upper row are the prediction accuracy of the five models.}
\label{fig:interpretation_cm_append}
\end{figure}

\begin{figure}[H]
\captionsetup[subfigure]{justification=centering}
\centering
\subfloat[PT ($69.2\%$)]{\includegraphics[width=0.2\linewidth]{exp/interpretation/pt_est_field_use_delta.png}\label{sfig:pt_est_field_append}}
\subfloat[PT-ResNet ($\delta = 10^{-5}$; $81.3\%$)]{\includegraphics[width=0.2\linewidth]{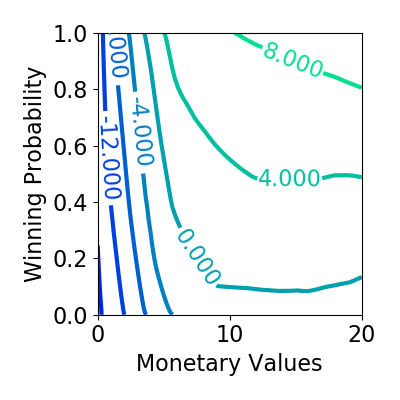}}
\subfloat[PT-ResNet ($\delta = 0.9$; $ 89.8\%$)]{\includegraphics[width=0.2\linewidth]{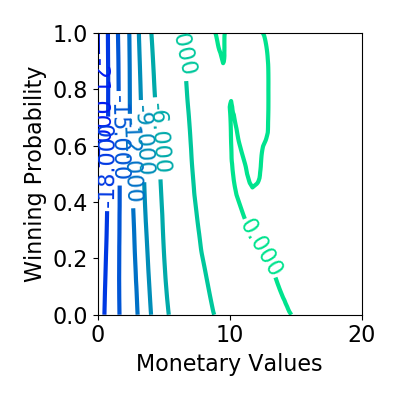}\label{sfig:pt_dnn_00001_field_append}}
\subfloat[PT-ResNet ($\delta = 0.99$; $90.0\%$)]{\includegraphics[width=0.2\linewidth]{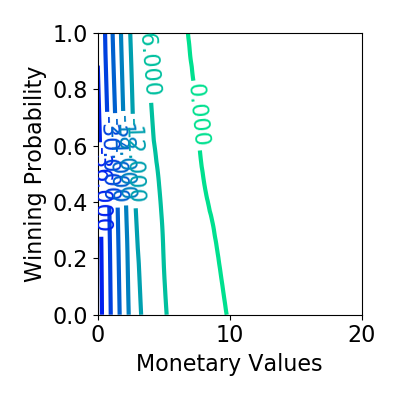}}
\subfloat[DNN ($89.0\%$)]{\includegraphics[width=0.2\linewidth]{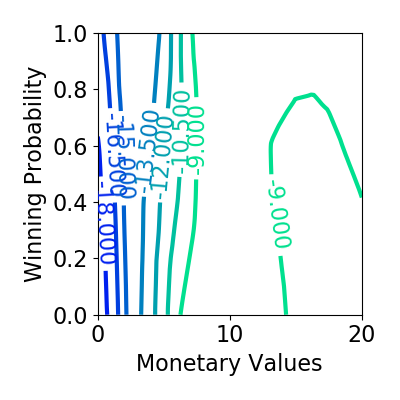}\label{sfig:pt_dnn_field_append}}
\\
\subfloat[Values]{\includegraphics[width=0.1\linewidth]{exp/interpretation/pt_est_x0_use_delta.png}\label{sfig:pt_est_x0_append}}
\subfloat[Prob]{\includegraphics[width=0.1\linewidth]{exp/interpretation/pt_est_x1_use_delta.png}\label{sfig:pt_est_x1_append}}
\subfloat[Values]{\includegraphics[width=0.1\linewidth]{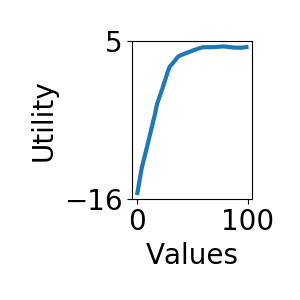}}
\subfloat[Prob]{\includegraphics[width=0.1\linewidth]{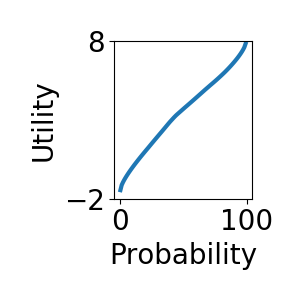}}
\subfloat[Values]{\includegraphics[width=0.1\linewidth]{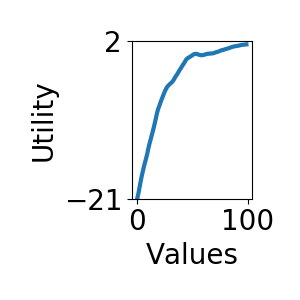}\label{sfig:pt_dnn_00001_x0_append}}
\subfloat[Prob]{\includegraphics[width=0.1\linewidth]{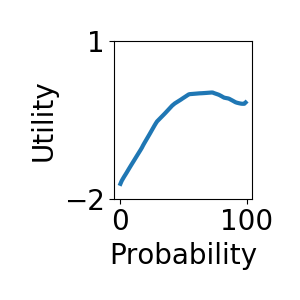}\label{sfig:pt_dnn_00001_x1_append}}
\subfloat[Values]{\includegraphics[width=0.1\linewidth]{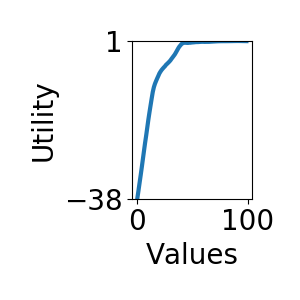}}
\subfloat[Prob]{\includegraphics[width=0.1\linewidth]{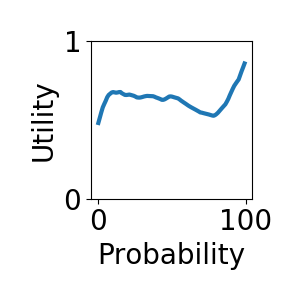}}
\subfloat[Values]{\includegraphics[width=0.1\linewidth]{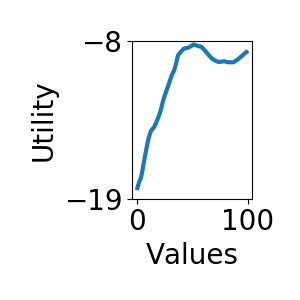}\label{sfig:pt_dnn_x0_append}}
\subfloat[Prob]{\includegraphics[width=0.1\linewidth]{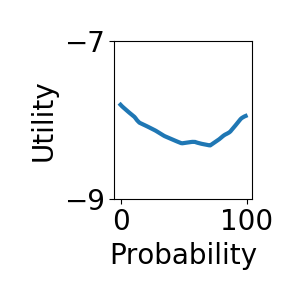}\label{sfig:pt_dnn_x1_append}}
\\
\caption{Utility functions of PT-ResNets, PT, and DNNs; upper row: visualization of 2D utility functions; lower row: visualization of 1D utility functions; percentages on the upper row are the prediction accuracy of the five models.}
\label{fig:interpretation_pt_append}
\end{figure}

\begin{figure}[H]
\captionsetup[subfigure]{justification=centering}
\centering
\subfloat[HD ($56.7\%$)]{\includegraphics[width=0.2\linewidth]{exp/interpretation/hd_est_field_use_delta.png}\label{sfig:hd_est_field_append}} 
\subfloat[HD Resnet ($\delta = 10^{-8}$; $58.4\%$)]{\includegraphics[width=0.2\linewidth]{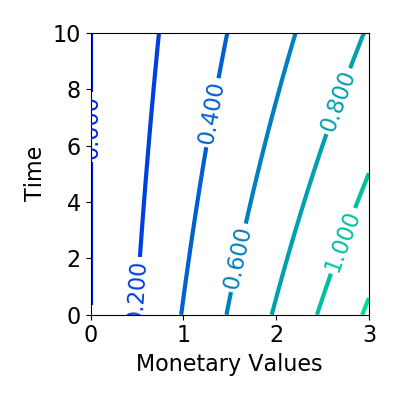}}
\subfloat[HD Resnet ($\delta = 0.05$; $75.4\%$)]{\includegraphics[width=0.2\linewidth]{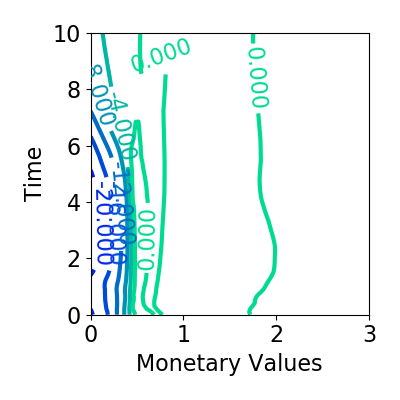}}
\subfloat[HD Resnet ($\delta = 0.99$; $74.9\%$)]{\includegraphics[width=0.2\linewidth]{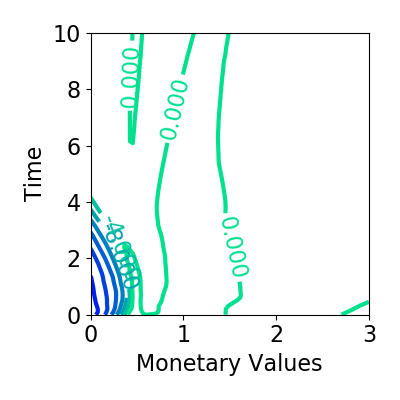}}
\subfloat[DNN ($72.6\%$)]{\includegraphics[width=0.2\linewidth]{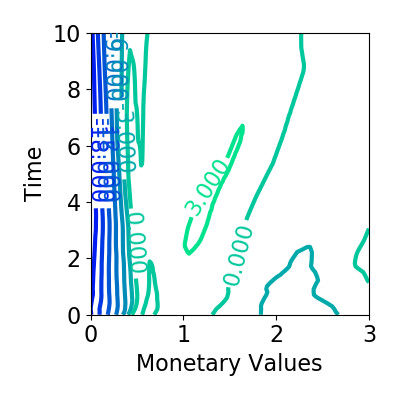}\label{sfig:hd_dnn_field_append}}\\
\subfloat[Values]{\includegraphics[width=0.1\linewidth]{exp/interpretation/hd_est_x0_use_delta.png}\label{sfig:hd_est_x0_append}}
\subfloat[Time]{\includegraphics[width=0.1\linewidth]{exp/interpretation/hd_est_x1_use_delta.png}\label{sfig:hd_est_x1_append}}
\subfloat[Values]{\includegraphics[width=0.1\linewidth]{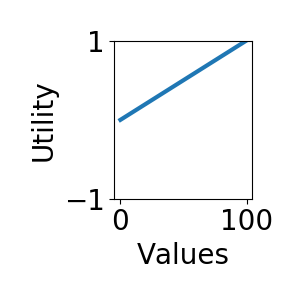}}
\subfloat[Time]{\includegraphics[width=0.1\linewidth]{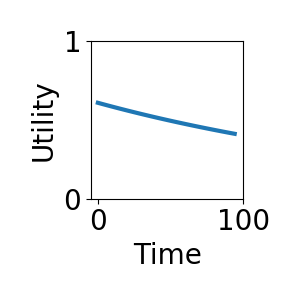}}
\subfloat[Values]{\includegraphics[width=0.1\linewidth]{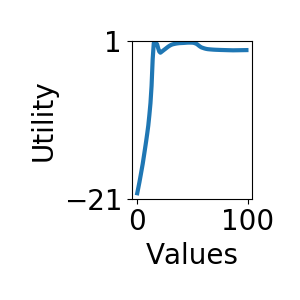}}
\subfloat[Time]{\includegraphics[width=0.1\linewidth]{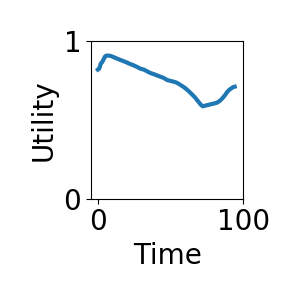}}
\subfloat[Values]{\includegraphics[width=0.1\linewidth]{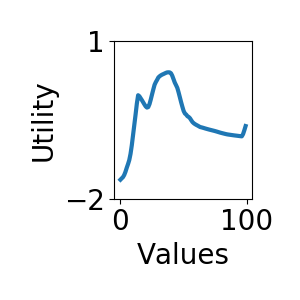}}
\subfloat[Time]{\includegraphics[width=0.1\linewidth]{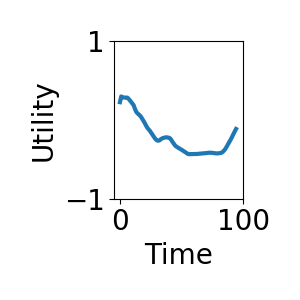}}
\subfloat[Values]{\includegraphics[width=0.1\linewidth]{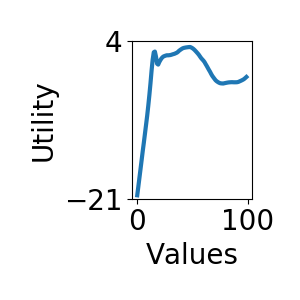}\label{sfig:hd_dnn_x0_append}}
\subfloat[Time]{\includegraphics[width=0.1\linewidth]{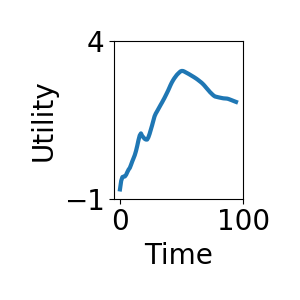}\label{sfig:hd_dnn_x1_append}}
\\
\caption{Utility functions of HD-ResNets, HD, and DNNs; upper row: visualization of 2D utility functions; lower row: visualization of 1D utility functions; percentages on the upper row are the prediction accuracy of the five models.}
\label{fig:interpretation_hd_append}
\end{figure}

\end{document}